\pdfoutput=1

\documentclass[11pt]{article}

\usepackage[]{acl}

\usepackage{times}
\usepackage{latexsym}

\usepackage[normalem]{ulem}
\usepackage{mathrsfs}
\usepackage{amsthm}
\usepackage{amsmath}
\usepackage{amssymb, bm}
\usepackage{multirow}
\usepackage{booktabs}
\usepackage{enumitem}
\usepackage{times}
\usepackage{latexsym}
\usepackage{array}
\usepackage{multirow}
\usepackage{graphicx}
\usepackage{amsfonts}
\usepackage{subcaption}
\usepackage{xspace}
\usepackage{tabularx}
\usepackage{algorithm}
\usepackage{arydshln}
\usepackage{algpseudocode}
\usepackage{colortbl}

\newcommand{\ourmodel}{PLAT\xspace}

\newcommand{\secref}[1]{\S\ref{#1}}

\algnewcommand\algorithmicinput{\textbf{Input:}}
\algnewcommand\algorithmicoutput{\textbf{Output:}}
\algnewcommand\Input{\item[\algorithmicinput]}
\algnewcommand\Output{\item[\algorithmicoutput]}

\definecolor{mypurple}{RGB}{111,61,121}
\definecolor{myblue}{RGB}{46,88,180}
\definecolor{myred}{RGB}{181,68,106}
\definecolor{textorange}{RGB}{237,125,49}
\definecolor{textblue}{RGB}{46,117,181}
\definecolor{textgreen}{RGB}{112,173,71}

\newcommand{\setblue}[1]{{\color{textblue}{#1}}}
\newcommand{\setgreen}[1]{{\color{textgreen}{#1}}}
\newcommand{\setred}[1]{{\color{black}{#1}}}

\newcommand{\setorange}[1]{{\color{textorange}{#1}}}
\newcommand{\setredd}[1]{{\color{myred}{#1}}}

\usepackage[T1]{fontenc}

\usepackage[utf8]{inputenc}

\usepackage{microtype}

\title{Phrase-level Textual Adversarial Attack with Label Preservation}

\author{Yibin Lei\textsuperscript{1}, Yu Cao\textsuperscript{2}, Dianqi Li\textsuperscript{3}, Tianyi Zhou\textsuperscript{3,4}, Meng Fang\textsuperscript{1}, Mykola Pechenizkiy\textsuperscript{1}\\
\textsuperscript{1}{Eindhoven University of Technology (TU/e)}\quad
\textsuperscript{2}{The University of Sydney}\\
\textsuperscript{3}{University of Washington}\quad
\textsuperscript{4}{University of Maryland}\\
{\tt y.lei2@student.tue.nl, ycao8647@uni.sydney.edu.au}\\ {\tt \{dianqili,tianyizh\}@uw.edu, \{m.fang,m.pechenizkiy\}@tue.nl}
}
\begin{document}
\maketitle
\begin{abstract}
Generating high-quality textual adversarial examples is critical for investigating the pitfalls of natural language processing (NLP) models and further promoting their robustness. 
Existing attacks are usually realized through word-level or sentence-level perturbations, which either limit the perturbation space or sacrifice fluency and textual quality, both affecting the attack effectiveness.
In this paper, we propose \textbf{P}hrase-\textbf{L}evel Textual \textbf{A}dversarial A{\textbf{T}}tack (\ourmodel) that generates adversarial samples through phrase-level perturbations. \ourmodel first extracts the vulnerable phrases as attack targets by a syntactic parser, and then perturbs them by a pre-trained blank-infilling model. Such flexible perturbation design substantially expands the search space for more effective attacks without introducing too many modifications, and meanwhile maintaining the textual fluency and grammaticality via contextualized generation using surrounding texts.
Moreover, we develop a label-preservation filter leveraging the likelihoods of language models fine-tuned on each class, rather than textual similarity, to rule out those perturbations that potentially alter the original class label for humans.
Extensive experiments and human evaluation demonstrate that \ourmodel has a superior attack effectiveness as well as a better label consistency than strong baselines.\footnote{Code is available at \url{https://github.com/Yibin-Lei/PLAT}}
\end{abstract}

\section{Introduction}\label{intro}
Despite the widespread success of deep learning in natural language processing (NLP) applications, 
a variety of works~\cite{universal_trigger,jia_certified,cheng2019robust} discovered that neural networks can be easily fooled to produce incorrect predictions, when their input text is modified by adversarial attacks that do not necessarily alter human predictions and the true meaning of the original text. 
Through the lens of adversarial attacks, we can allocate the weakness of models and in turn improve their reliability and robustness~\cite{jia2017adversarial,belinkov2018_mt}.


However, generating high-quality adversarial texts is nontrivial due to the discrete nature of human language and its rigorous linguistic structures. 
While many efforts of previous works have been taken to generate word-level perturbations~\cite{ren_pwws,alzantot_generating,jin_textfooler,li_bert_attack,garg_bae,li_clare} for the sake of simplicity,
their attacks are restricted to independent perturbations on single words and thus cannot produce richer and more diverse forms of adversarial examples. 
To expand the search space for attacks, sentence-level attacks have been explored~\cite{iyyer_paraphrase,wang_cat_gan,wang_t3, qi_style} such as using paraphrasing, but their textual quality is usually poor due to insufficient constraints or controls on the structure and meanings of the generated texts. 
\begin{figure}
\setlength{\abovecaptionskip}{0.1cm}
\setlength{\belowcaptionskip}{-0.2cm}
    \centering
    \includegraphics[width=\linewidth]{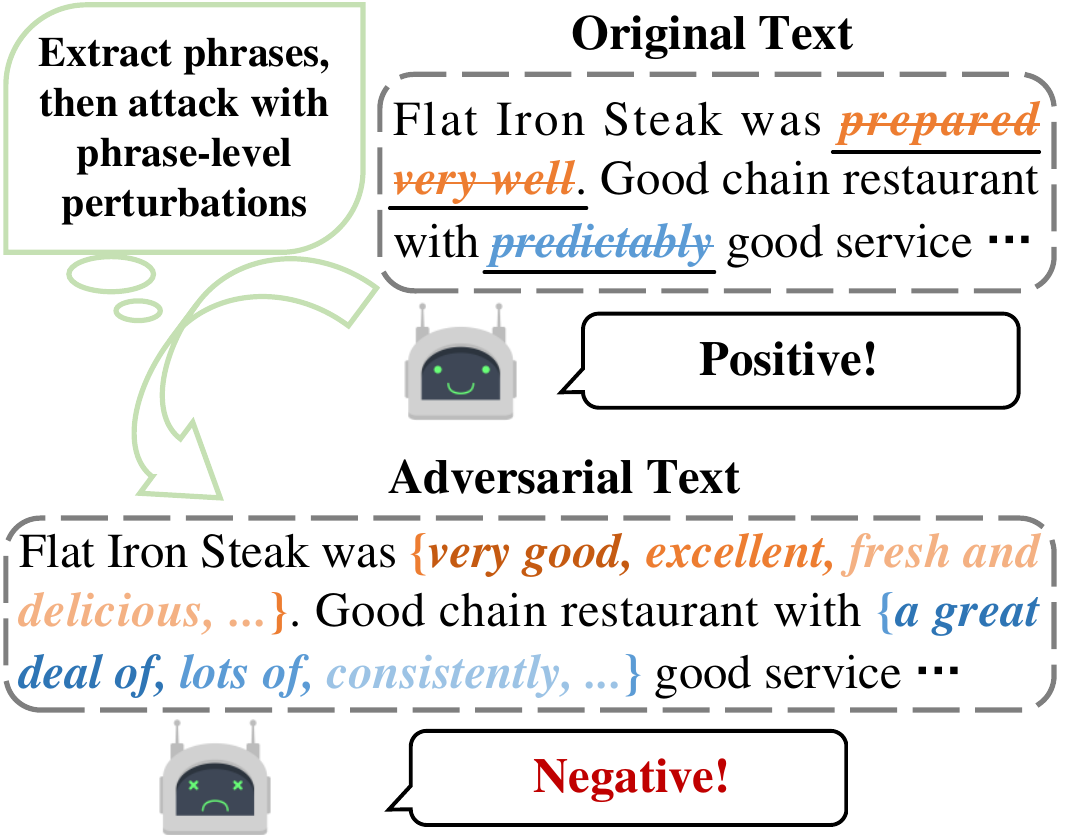}
    \caption{In \ourmodel, we extract phrases from the original text as attack targets, then use a blank-infilling model to obtain perturbation candidates and generate effective adversarial texts. Note that both target phrases and perturbations may contain one or multiple tokens.}
    \label{fig:illustration}
\end{figure}

To generate controllable high-quality textual adversarial examples, we propose a new phrase-level attack, \ourmodel, which can explore more diverse and flexible forms of perturbations than single word perturbations. Our model is able to produce phrase-level perturbations with a high success rate and preserve the textual similarity in a more controllable manner.
As illustrated in Figure~\ref{fig:illustration},  with the help of constituency parsing, \ourmodel first detects and extracts the most vulnerable phrases from the text to the victim model as the attack targets. To maintain textual fluency and grammaticality, \ourmodel perturbs these phrases through a contextualized blank-infilling procedure by a pre-trained language model. 
Compared to existing textual adversarial attacks, \ourmodel can produce more effective attacks by searching in a larger space of phrase-level perturbation. Meanwhile, \ourmodel delicately controls the range and amount of modifications so the textual meaning of the original texts will not be altered significantly after attacks.

Moreover, the success of attacks can be trivial if allowed to arbitrarily distort the ground-truth label or key contents. 
Hence, a valid attack is required to not change the ground-truth label predicted by humans. 
However, the semantic similarity filters widely used in existing works \setred{to guarantee the validity of adversarial samples}~\cite{jin_textfooler,li_bert_attack} perform unsatisfactorily\setred{, especially} in preserving the textual meaning and even flip the gold labels according to a recent study~\cite{morris_reevaluating}.  
To this end, we develop a label-preservation filter to maintain class-dependent properties such as sentiments. It is built upon the comparison of likelihoods of language models \setred{fine-tuned} on different classes' data. Thereby, it selects the attacks that can easily fool the victim model but \setred{possibly} hardly alter the original labels.

Our contributions in this paper are three-fold:
\begin{itemize}[wide=0\parindent]
    \item We propose a phrase-level textual adversarial attack that employs contextualized blank-infilling to generate high-quality phrase perturbations. It expands the perturbation space of word-level attacks and thus can produce more effective attacks without notably hurting the fluency and grammaticality.
    \item We introduce a novel label-preservation filter, which utilizes the likelihood of class-specific language model to generate more reliable adversarial examples. 
    \item Extensive experiments demonstrate the effectiveness of \ourmodel on multiple text classification and natural language inference tasks, hence presenting a new robustness challenge to existing NLP models.\looseness-1 
\end{itemize}

\section{Methodology}

In this section, we first formulate the problem of phrase-level textual adversarial attack. Then we elaborate on how \ourmodel chooses phrase candidates to attack and how to adversarially perturb them. Finally, we discuss strategies to select the most effective perturbations with label preservation.

\subsection{Problem Definition}

We focus on generating textual adversarial examples for classification tasks. Given a textual sequence $\mathbf{x}=x_1x_2 \ldots x_n$ with a specific attribute label $y$ and a victim model $F$ (assume $F(\mathbf{x})=y$) to attack, our goal is to generate an adversarial sample $\mathbf{x}'$ by perturbing $\mathbf{x}$. 
A valid adversarial example $\mathbf{x}'$ can successfully trigger a wrong prediction of the victim model, i.e. $F(\mathbf{x}') \neq y$, while the human judgement on $\mathbf{x}'$ should stay unaltered as $y$. To achieve this goal, $\mathbf{x}'$ needs to be sufficiently similar to $\mathbf{x}$ with reasonable fluency and correct grammaticality. 

\subsection{Phrase-level Attack}
\paragraph{Phrase candidates.}
Given a sequence $\mathbf{x}$, \ourmodel allocates candidates of phrases to attack from the syntactic tree extracted by a language parser (e.g., Stanford Parser~\cite{manning_stanford}, etc.). The model first traverses all constituents (nodes) in the syntactic tree in a top-down manner. If a node is identified as a phrase, i.e. tagged as \emph{NP}, \emph{VP}, etc., the text piece in $\mathbf{x}$ associated with all nodes in the subtree that is rooted at the current node, will be regarded as an attacking candidate. For more controllable attacks, we set a maximum depth of syntactic subtree $d$ to restrict the length of candidate phrases so the modification to $\mathbf{x}$ is limited, hence resulting in more \setred{possibly} valid adversarial samples.
Thereby, \setred{we obtain} a set of \setred{multiple} candidate phrases in the form of $\mathcal{A}=\{(\mathbf{a},i, j)\}$, where $i$ and $j$ are the indices of the leftest and rightest token of a phrase $\mathbf{a}$ from $\mathbf{x}$.

\paragraph{Phrase importance.}
To produce \setred{attacks more effectively} against the victim model $F$ (e.g., \setred{fine-tuned} BERT~\cite{devlin_bert}), \ourmodel only perturbs the phrase candidates important to the prediction \setred{as former works}~\cite{jin_textfooler,ren_pwws}.
Specifically, we consider to replace a phrase $\{(\mathbf{a},i, j)\}$ in $\mathbf{x}$ with a series of special symbol \text{[MASK]}\footnote{We empirically found this is better than  single-mask replacement.} with the same length as $\mathbf{a}$, which results in $\tilde{\mathbf{x}} = x_1 \ldots x_{i-1}, \text{[MASK]} \ldots \text{[MASK]}, x_{j+1} \ldots x_n$.
The importance for phrase $\mathbf{a}$ is measured by
\begin{equation*}
\label{eq:importance}
    I(\mathbf{a})=P_F(y \ | \ \mathbf{x}) - P_F(y \ | \ \tilde{\mathbf{x}}),
\end{equation*}
where $P_F(y|\cdot)$ is the probability of the ground truth label $y$ predicted by $F$ given the input \setred{textual sequence}. Larger $I(\mathbf{a})$ indicates that the phrase $\mathbf{a}$ has more significant contribution to the prediction of $y$. \ourmodel manipulates each \emph{target phrase} in candidate sets $\mathcal{A}$ by following \setred{a} descending order of their importance scores. So more effective phrase-level perturbations are applied earlier for achieving minimum modifications to $\mathbf{x}$.

\paragraph{Phrase perturbations.}
To generate phrase-level adversarial perturbations, \ourmodel performs a blank-infilling procedure on each target phrase.
Specifically, \ourmodel first replaces a target phrase $\mathbf{a}\in\mathcal{A}$ with a blank \setred{(a single special [MASK] token in our implementation)} from index $i$ to $j$, i.e., 
\begin{equation}
    \nonumber
        \tilde{\mathbf{x}}_{\backslash \mathbf{a}} = x_1,  \ldots, x_{i-1}, \underline{\quad\quad\quad}, x_{j+1}, \ldots, x_n.
\end{equation}
Then a pre-trained blank-infilling language model, e.g., BART~\cite{lewis_bart} or T5~\cite{raffel_t5}, takes $\tilde{\mathbf{x}}_{\backslash \mathbf{a}}$ as the input and fills a phrase $\mathbf{b}=z_1 \ldots z_m$ \setred{having $m$ tokens} into the blank conditioned on surrounding context, i.e., 
\begin{equation}
\nonumber
\begin{split}
    \tilde{\mathbf{x}}_{\mathbf{b}} = x_1 \ldots x_{i-1}, z_1 \ldots z_m, x_{j+1} \ldots x_n.
\end{split}
\end{equation}
In contrast to paraphrasing each phrase independently, such contextualized infilling procedure can produce more fluent and grammatically correct perturbations fitting into the rest context. 

For attacking each target phrase, \ourmodel samples $N$ candidates of perturbed phrases $\mathcal{B}=\{\mathbf{b}\}$ with varying lengths. During the generation, \ourmodel tends to sample tokens of higher probability at every step so that the outputs are \setred{more appropriate} with the surrounding context. 
\setred{In addition,} we keep the maximum length of perturbations not greater than the length of original phrases plus a threshold $l$ (e.g., $|\mathbf{b}|\leq|\mathbf{a}|+l$). The most effective perturbation in $\mathcal{B}$ is then selected to replace the target phrase $\mathbf{a}$, resulting in a perturbed text $\tilde{\mathbf{x}}_{\mathbf{b}}$ \setred{(The details of selection will be discussed in \secref{sec:label_scoring}).}


We apply the above phrase perturbation sequentially to all target phrases from $\mathcal{A}$\footnote{If a phrase $\mathbf{b}$ is perturbed, phrases that overlap $\mathbf{b}$ in the remaining phrases of $\mathcal{A}$ will be ignored.} until (1) a valid adversarial sample $\mathbf{x}^{(t)}$ is found when perturbing the $t^{th}$ target phrase, i.e., $F(\mathbf{x}^{(t)}) \neq y$;
or (2) the maximum number of perturbations $T$ is reached. We summarize the above procedure in Algorithm~\ref{alg:phrase_attack}.

\begin{algorithm}[t]
\centering
\caption{Adversarial Attack by \ourmodel }
\label{alg:phrase_attack}
\begin{algorithmic}[1]
    \State \textbf{Input:} Original text $\mathbf{x}$, the gold label $y$, victim model $F$, maximum number of perturbation $T$, importance score $I$, \setred{likelihood ratio threshold $\delta$}. \setred{Two filter functions: class-conditioned likelihood function $R(\cdot)$, effectiveness score function $S(\cdot)$}
    \State \textbf{Output:} An adversarial example $\mathbf{x}'$
    \State Extract phrase candidates from $\mathbf{x}$ to form the set $\mathcal{A}$
    \State $\mathbf{x}^{(0)} \gets \mathbf{x}$
    \For{$1 \leq t \leq T$}
        \State $\mathbf{a} \gets $ target phrase with highest $I$ in $\mathcal{A}$
        \State $\mathcal{B} \gets $ a set of phrases perturbations generated by blank-infilling $\tilde{\mathbf{x}}_{\backslash \mathbf{a}}^{(t-1)}$
        \State $\mathcal{B} \gets $ filtering $\mathcal{B}$ by $R(\mathbf{x}^{(t-1)}, \mathbf{b'}, y) < \delta$
        \If{$\mathcal{B} = \varnothing$} $\mathbf{x}^{(t)} = \mathbf{x}^{(t-1)}$, \textbf{continue}
        \EndIf
        \State $\mathbf{b} \gets \mathop {{\rm{argmax}}}\limits_{\mathbf{b'} \in \mathcal{B}} S(\mathbf{x}^{(t-1)},\mathbf{b'})$ 
        \State $\mathbf{x}^{(t)} \gets \tilde{\mathbf{x}}^{(t-1)}_{\mathbf{b}}$(replace $\mathbf{a}$ with $\mathbf{b}$ in $\mathbf{x}^{(t-1)}$)
        \If{$F(\mathbf{x}^{(t)}) \neq y$} \textbf{return} $\mathbf{x}^{(t)}$
        \EndIf
    \EndFor
    \State \textbf{return} \textsc{None}
\end{algorithmic}
\end{algorithm}

\subsection{Label Preservation and Effective Perturbation.} 
\label{sec:label_scoring}
\paragraph{Label preservation filter.}

Although existing works~\cite{jin_textfooler,chen_maya} usually employ a semantic similarity constraint (e.g., USE~\cite{cer_use}) to encourage the validity of adversarial samples, it has been observed that such constraint is unreliable to preserve the textual meaning~\cite{morris_reevaluating}. Moreover, existing approaches rarely preserve class-dependent contents, e.g., sentiments, and might produce invalid adversarial examples with human-\setred{annotated} labels flipped. Such a drawback is commonly observed in our human evaluation in~\secref{sec:exp_res}.

In order to retain the class-related characteristics most critical to  classification tasks, 
inspired by~\citealp{malmi_unsupervised_style}, \ourmodel directly filters phrase perturbations using likelihoods provided by class-conditioned masked language models (CMLMs). 
Specifically, given a sequence $\tilde{\mathbf{x}}_{\mathbf{b}} = x_1,  \ldots, x_{i-1}, z_1,\ldots$ $, z_m, x_{j+1}, \ldots, x_n$, the  class-conditioned likelihood of the adversarially perturbed phrase $\mathbf{b}=z_1, \ldots, z_m$ for phrase $(\mathbf{a},i,j)$ in $\mathbf{x}$ can be calculated as
\begin{equation*}
\label{eq:likelihood}
    L(\mathbf{x}, \mathbf{b}, y) = \prod_{k=1}^{m}  P_{\mathrm{CMLM}}\left({z}_{k} \mid \boldsymbol{\tilde{\mathbf{x}}}_{\mathbf{b} \backslash z_k} ; \Theta_\textit{y}\right).
\end{equation*}
Here, $m$ is the length of $\mathbf{b}$, $\tilde{\mathbf{x}}_{\mathbf{b} \backslash z_k}$ is \setred{the perturbed sequence} $\tilde{\mathbf{x}}_{\mathbf{b}}$ with token $z_k$ masked, $P_{\mathrm{CMLM}}$ is the likelihood of $z_k$ given $\tilde{\mathbf{x}}_{\mathbf{b} \backslash z_k}$, which is produced by a class-conditioned masked language model $\Theta_\textit{y}$ conditioned on class $y$. 
The conditional language model is first initialized as a pre-trained model and then fine-tuned with the pre-training objective on the data \setred{from the dataset that} belonging to the class $y$.
Therefore, a larger likelihood indicates that $\mathbf{b}$ is more likely to match the corresponding class distribution given the surrounding context.\footnote{In practice, we partition the whole text into multiple sentences and the likelihood $P_{\mathrm{CMLM}}$ for a phrase $\mathbf{b}$ is calculated locally using its corresponding sentence.}


To avoid label flipping of human prediction, the phrase perturbations should enjoy a higher likelihood on the original class's distribution but a lower likelihood on other classes. 
This property can be measured by the following likelihood ratio:
\begin{equation*}
\label{eq:norm_likelihood}
    R(\mathbf{x}, \mathbf{b}, y) = L(\mathbf{x}, \mathbf{b}, y) / \mathop {\rm{max}}\limits_{\tilde{y} \in \mathcal{Y}, \tilde{y} \neq y} L(\mathbf{x}, \mathbf{b}, \tilde{y}),
\end{equation*}
where $\mathcal{Y}$ denotes the set of all classes in the task. A higher likelihood ratio suggests the perturbation is more correlated to the original label in contrast to other labels. For better label preservation, the committed phrase perturbations are required to have a likelihood ratio larger than certain threshold $\delta$, i.e. $R(\mathbf{x}, \mathbf{b}, y) \geq \delta$ for $\mathbf{b}\in\mathcal{B}$. As shown in~\secref{sec:exp_res}, our method outperforms other baselines on label-preservation \setred{according to human evaluation}.


\paragraph{Selection of the most effective perturbation.}

To generate $\mathbf{x}'$ with sufficient global textual-similarity to $\mathbf{x}$, \ourmodel selects target phrases \setred{with syntactic tree depth no deeper than $d$} and restricts their perturbations' lengths to be \setred{no larger than their length by $l$} . Moreover, \ourmodel aims at utilizing minimum perturbations to perform effective adversarial attacks, so the textual similarity can be preserved. Minimum perturbations can in return help maintain reasonable fluency and grammaticality of the generated texts.

To achieve the above goals, \ourmodel selects the most effective phrase perturbation at each step as the one that minimizes the probability of the gold label $y$ predicted by $F$. We use a score to measure each phrase $\mathbf{b}$ in terms of how likely it can successfully fool the model, i.e. the negative probability of the gold label $y$ for the original $\mathbf{x}$ associated with the perturbation  $\mathbf{b}$, i.e.,
\begin{equation*}
\label{eq:score}
    S(\mathbf{x}, \mathbf{b}) = -P_F(y \ | \ \tilde{\mathbf{x}}_{\mathbf{b}}).
\end{equation*}
When attacking a target phrase, \ourmodel only chooses one phrase perturbation $\mathbf{b}\in\mathcal{B}$ with the highest score. The resulted perturbed-sequence is retained and then used as the initial sequence for the next time of perturbation.

\subsection{Discussion}
A primary novelty of \ourmodel is the phrase-level perturbation. Compared to the widely studied word-level perturbations~\cite{ren_pwws,jin_textfooler,li_clare} that can only independently perturb a single word every time, \ourmodel can perturb a text span of varying lengths by replacing it with phrases of possibly unequal lengths. Hence, it produces a more flexible attack by searching it in a larger perturbation space. Although the textual phrase-level attack has been studied by a concurrent work MAYA~\cite{chen_maya}, there are several critical differences of \ourmodel, i.e.,
\begin{enumerate}[wide=0\parindent, noitemsep, topsep=0pt]
    \item[(1)] The phrase-level attack by \ourmodel is a more general attack model that covers both word-level and phrase-level perturbations in one framework, while MAYA builds separate sub-modules for different levels of perturbations. 
    \item[(2)] \ourmodel adopts a blank-infilling strategy and leverages language models to generate phrase perturbations in a context-aware manner, leading to more fluent and grammatical adversarial examples. On the contrary, MAYA applies paraphrasing to each constituent target separately without taking its surrounding context information into account. 
    \item[(3)] \ourmodel applies several constraints and filters to the phrase perturbations for more controllable attacks and better preservation of the original textual and label information, while MAYA has no such restrictions and its generated perturbations can introduce arbitrary distortions to the original text. 
\end{enumerate}

\section{Experiments}
In this section, we first elaborate on the experimental settings and implementation details of \ourmodel as well as the comparisons to several baselines in \secref{sec:setup}. We then introduce the datasets and evaluation designs in \secref{sec:data}. At the end, we summarize the main results in \S\ref{sec:exp_res}.

\subsection{Setup}\label{sec:setup}
The implementation details are given as follows:
\begin{itemize}[wide=0\parindent, noitemsep, topsep=0pt]
    \item We use pre-trained BART$_{\text{base}}$~\cite{lewis_bart} as the language model for blank-infilling to generate phrase perturbations. We sample $N=5000$ candidates \setred{as the phrase set $\mathcal{B}$ via} Top-K sampling~\cite{fan2018hierarchical}, while set \setred{restriction} $d=4$, $l=3$ for each target phrase. In sections~\secref{sec:ablation}, we also report the performance \setred{comparison} when using different language models.
    \item We use \setred{a} RoBERTa$_{\text{base}}$~\cite{liu_roberta} to calculate the class-conditioned likelihood for label-preserving filters. On each dataset, the model is further fine-tuned to optimize the \setred{masked language modeling} pre-training objective on each sequence with a prepending special label token \setred{indicating the current class}. We set threshold $\delta=1$ for the filtering.
    \item The victim model $F$ is an MLP classifier based on  \setred{a BERT$_{\text{base}}$ model~\cite{devlin_bert}}. It takes the representation of [CLS] token for prediction and is fine-tuned on the target datasets in advance.
\end{itemize}

\paragraph{Baselines.}
We compare \ourmodel with \setred{four} state-of-the-art textual adversarial attack models:\footnote{All results are obtained by running their released code.}
\begin{itemize}[wide=0\parindent, noitemsep, topsep=0pt]
    \item
    \textbf{Textfooler}~\cite{jin_textfooler}: 
    a word-level attack model, which replaces tokens with their synonyms via counter-fitting word embeddings~\cite{mrkvsic_counter_fitting}. USE~\cite{cer_use} distance is used to select adversarial texts \setred{that can mostly} preserve the semantic similarity.
    \item
    \textbf{CLARE}~\cite{li_clare}: 
    instead of token replacement only, \textbf{CLARE} considers three word-level perturbations\setred{: replacement, insertion, and merging}. Pre-trained masked language models are used to generate perturbations and a USE semantic similarity filter is applied.
    \item
    \textbf{MAYA}~\cite{chen_maya}: 
    a multi-granularity model that attacks the input using two separate modules for word replacement and constituent paraphrasing. It employs the embedding of Sentence-BERT~\cite{reimers_sentence_bert} for semantic similarity preservation.
    \item
    \setred{\textbf{StyleAdv}~\cite{qi_style}:
    a sentence-level model based on text style transfer. To launch an attack, it paraphrases the whole examples with five different text styles: tweets, bible, poetry, shaekspeare and lyrics.}
\end{itemize}

\begin{table}[t]
\setlength{\abovecaptionskip}{0.1cm}
\setlength{\belowcaptionskip}{-0.2cm}
\centering
\setlength{\tabcolsep}{4.0pt}
\small{
\begin{tabular}{@{} lccrrr @{}}
\toprule
\textbf{Dataset}  &
\textbf{Avg. Len} & \textbf{\#Classes} & \textbf{Train} & \textbf{Test} & \textbf{Acc} 
\\\midrule

Yelp & 130 &  2 & 560k & 38k & 91.8\%\\
AG News & 46 & 4 & 120k   & 7.6k  & 94.6\%\\
\midrule

MNLI & 23/11 & 3  & 392k & 9.8k & 84.0\%   \\
QNLI & 11/31  & 2 & 105k & 5.4k & 91.4\%    \\
\bottomrule
\end{tabular}
}
\caption{Statistics of datasets and the performance of victim models on each dataset.}
\label{tab:dataset}
\end{table}

\begin{table*}[t!]
\setlength{\abovecaptionskip}{0.1cm}
\setlength{\belowcaptionskip}{-0.3cm}
\small
\centering
\setlength{\tabcolsep}{2mm}

\begin{tabular}{lrrrrrrrrrrr}
\toprule
\textbf{Dataset} & \multicolumn{5}{c}{Yelp (PPL = 51.5)} & \ \ & \multicolumn{5}{c}{AG News (PPL = 62.8)} \\
\midrule
\textbf{Model} & \textbf{ASR$\uparrow$} & \textbf{DIS$\downarrow$} & \textbf{BLEU$\uparrow$}& \textbf{PPL$\downarrow$} & \textbf{GER$\downarrow$} & \ \ &  \textbf{ASR$\uparrow$} & \textbf{DIS$\downarrow$} & \textbf{BLEU$\uparrow$}& \textbf{PPL$\downarrow$} & \textbf{GER$\downarrow$}  \\
\midrule
Textfooler & 94.5 & 0.11 & 0.80 & 101.1 & 0.73 & \ \ & 65.5 & 0.29  & 0.52 & 339.0 & 1.43 \\
CLARE & 97.3 & \textbf{0.07} & \textbf{0.88} & 65.2 & {0.08} & \ \ & 68.0 & \textbf{0.09} & \textbf{0.86} & 97.2 & -0.03 \\
\arrayrulecolor{black!50}\specialrule{.3pt}{1pt}{1pt}
MAYA & 97.0 & 0.43 & 0.44 & 78.9 & 5.23 & \ \ & 94.2 & 0.64 & 0.25 & 168.6 & 4.30 \\
\setred{StyleAdv} & 90.1 & 0.96 & /  & 132.0 & \textbf{-0.64} & \ \ & 79.6 & 0.97 & / & 111.3 & \textbf{-0.31} \\
\ourmodel & \textbf{98.4} & 0.17 & 0.78 & \textbf{56.8} & 0.33 & \ \ & \textbf{95.7} & 0.34 & 0.58 & \textbf{80.3} & 0.58 \\
\bottomrule
\end{tabular}

\begin{tabular}{lrrrrrrrrrrr}
\toprule
\textbf{Dataset} & \multicolumn{5}{c}{MNLI (PPL = 60.9)} & \ \ & \multicolumn{5}{c}{QNLI (PPL = 46.0)} \\
\midrule
\textbf{Model} & \textbf{ASR$\uparrow$} & \textbf{DIS$\downarrow$} & \textbf{BLEU$\uparrow$}& \textbf{PPL$\downarrow$} & \textbf{GER$\downarrow$} & \ \ &  \textbf{ASR$\uparrow$} & \textbf{DIS$\downarrow$} & \textbf{BLEU$\uparrow$}& \textbf{PPL$\downarrow$} & \textbf{GER$\downarrow$}  \\
\midrule
Textfooler & 58.6 & 0.15 & 0.71 & 159.0 & 0.67 & \ \ & 57.8 & 0.19 & 0.63 & 164.5 & 0.62 \\
CLARE & 86.2 & \textbf{0.10} & \textbf{0.82} & 82.7 & 0.09 & \ \ & 82.6 & \textbf{0.15} & \textbf{0.74} & 75.9 & \textbf{0.03} \\
\arrayrulecolor{black!50}\specialrule{.3pt}{1pt}{1pt}
MAYA & 92.8 & 0.40 & 0.49 & 104.7 & 2.20 & \ \ & 78.6 & 0.40 & 0.48 & 101.4 & 2.90 \\
\setred{StyleAdv} & 72.5 & 0.96 & /  & 103.7 & \textbf{-0.53} & \ \ & 67.3 & 0.95 & / & 108.3 & 0.17 \\
\ourmodel & \textbf{96.6} & 0.20 & 0.74 & \textbf{62.1} & 0.23 & \ \ & \textbf{92.4} & 0.25 & 0.68 & \textbf{51.3} & 0.06 \\
\bottomrule
\end{tabular}
\caption{\label{tab:main_results}Adversarial attack performance of \ourmodel and baselines on four datasets, in terms of attack success rate (ASR), edit distance (DIS), BLEU, perplexity (PPL), and increased grammar errors (GER). \textbf{Bold values} indicate the best performance for each metric. $\downarrow$($\uparrow$) indicates the higher (lower) the better. Note that phrase-level attacks naturally introduce more modifications than word-level attacks so they are not directly comparable on DIS and BLEU metrics in the table.}
\end{table*}

\subsection{Datasets and Evaluation}\label{sec:data}
\paragraph{Datasets.} We investigate the following datasets for text classification and natural language inference tasks in our experiments. The statistics and performance of the victim models evaluated on each dataset are reported in Table~\ref{tab:dataset}.
\begin{itemize}[wide=0\parindent, noitemsep, topsep=0pt]
\item \textbf{Yelp Reviews}~\citep{zhang2016characterlevel}: a binary sentiment classification dataset containing restaurant reviews as samples.

\item \textbf{AG News}~\citep{zhang2016characterlevel}: a news articles classification dataset covering four classes: \emph{World}, \emph{Sport}, \emph{Business}, and \emph{Science and Technology}.

\item \textbf{MNLI}~\citep{williams2018broad}: a natural language inference dataset, where each sample contains a pair of sentences whose relationship is labeled as \emph{entailment}, \emph{neutral}, or \emph{contradiction}. We use the \emph{matched} test set here.

\item \textbf{QNLI}~\citep{wang_glue}: a natural language inference dataset based on the question answering corpus SQuAD~\cite{rajpurkar_squad}. Each sample contains a context and a question labeled as \emph{entailed} or \emph{not entailed}.
\end{itemize}
All attacks will be conducted on 1000 instances randomly drawn from test sets. For tasks on a pair of sentences, we attack the longer sentence.

\paragraph{Evaluation metrics.} We evaluate models using the following automatic metrics:
\begin{itemize}[wide=0\parindent, noitemsep, topsep=0pt]
\item {\bf Attack success rate (ASR)}: the percentage of successful adversarial attacks that trigger wrong predictions of the victim model.
\item {\bf Edit Distance (DIS)}~\cite{navarro_edit_distrance}: the normalized Levenshtein distance that measures the minimum amount of word edits required to transform the original text to the adversarial one. It measures the modification rate of an adversarial sample.
\item {\bf BLEU}~\cite{papineni_bleu}: the BLEU score between an adversarial sample and its corresponding original sample is used to measure their n-gram overlap (textual similarity). \setred{We do not report the BLEU scores of \textbf{StyleAdv}, as the BLEU scores are extremely low (on 3 of 4 tasks the BLEU scores are at the scale of $10^{-81}$).}
\item {\bf Perplexity (PPL)}: a pre-trained GPT2$_{\text{small}}$~\cite{radford_gpt2} is used to calculate the PPL of adversarial texts, which reflects the fluency as suggested by~\cite{kann_sentence,zang2020word}.
\item {\bf Grammar error (GER)}: Following~\citealp{zang_word_grammar_error}, we employ LanguageTool\footnote{\url{https://www.languagetool.org/}} to calculate the average number of grammar errors newly introduced by adversarial samples. 
\end{itemize}
We only evaluate the last four metrics on the successful attacks against the victim model. 




\subsection{Main Results}\label{sec:exp_res}
Table~\ref{tab:main_results} summarizes the main experimental and comparison results. Overall, \ourmodel consistently achieves \setred{a} better attack success rate and perplexity performance across all datasets. We attribute this to the flexible phrase-level perturbations generated using  contextual information. \setred{\ourmodel achieves a significantly better text quality than the sentence-level model StyleAdv, in terms of modificaton rates, BLEU, and perplexity.} Compared with a phrase-level model MAYA, our method is significantly better on modification rates, BLEU, and grammar scores. Hence, despite not using semantic similarity constraints, \ourmodel is more controllable than MAYA as we confine the modification ranges and generate perturbations by contextual blank-infilling. While word-level attacks naturally introduce fewer perturbations and thus have better textual similarity and grammaticality, its perturbation space is small and results in lower success rates. On the contrary, \ourmodel achieves the highest success rate while \setred{on a par with} word-level attacks on textual similarity and grammaticality, hence achieving a sweet spot among all metrics.

\begin{table}[t!]
\centering

\small
\setlength{\tabcolsep}{4pt}
\setlength{\abovecaptionskip}{0.1cm}
\setlength{\belowcaptionskip}{-0.3cm}
\begin{tabular}{@{} l ccc @{}}
\toprule
Metric & \ourmodel & equal & CLARE
\\
\midrule
Meaning preservation & 39.8 & $\backslash$ & 33.3 \\
Label preservation & \textbf{77.1} & $\backslash$  & 49.8 \\
Fluency and grammaticality &  33.1 & 32.3 & 34.6 \\
\bottomrule
\end{tabular}
\setlength{\tabcolsep}{4.5pt}
\begin{tabular}{@{} l ccc @{}}
\toprule
Metric & \ourmodel & equal & MAYA
\\
\midrule
Meaning preservation & 39.8 & $\backslash$ & 30.2 \\
Label preservation & \textbf{77.1} & $\backslash$  & 53.5 \\
Fluency and grammaticality &  \textbf{45.0} & 29.0 & 26.0 \\
\bottomrule

\end{tabular}
\caption{Human evaluation performance in percentage on the Yelp dataset. 
}
\label{tab:human_evaluation_results}
\end{table}


\begin{table*}[t!]
\small
\setlength{\abovecaptionskip}{0.1cm}
\setlength{\belowcaptionskip}{-0.2cm}
\centering
\begin{tabularx}{\textwidth}{lX}
\toprule
\textbf{Yelp (Negative)} &  The quality of the food has really plummeted over the past year. We use to love coming her to get the creamy clam chowder, not its watery and gross.\\
\noalign{\vskip 2pt}\hdashline\noalign{\vskip 2pt}
\textbf{TextFooler (Positive)} & The quality of the food has really \setblue{\textbf{engulfed}} over the past year. We use to love coming her to get the creamy clam chowder, not its watery and gross.\\
\noalign{\vskip 2pt}\hdashline\noalign{\vskip 2pt}
\textbf{CLARE (Positive)} &  The quality of the food has really \setgreen{\textbf{soared}} over the past year. We use to love coming her to get the creamy clam chowder, not its watery and gross. \\
\noalign{\vskip 2pt}\hdashline\noalign{\vskip 2pt}
\textbf{MAYA (Positive)} & The quality of the food has really \setred{\setredd{last year was a big one for the fall}}. We use to love coming her to get the creamy clam chowder, not its watery and gross. \\
\noalign{\vskip 2pt}\hdashline\noalign{\vskip 2pt}
\textbf{\ourmodel (Positive)} & The quality of the food has \setorange{\textbf{also somewhat}} plummeted over the past year. We use to love coming her to get the creamy clam chowder, not its watery and gross.\\
\bottomrule
\end{tabularx}
\caption{\label{tab:case1} Adversarial examples generated by different models on Yelp dataset, perturbations are colored.}
\end{table*}

\paragraph{Human evaluation.}
We further conduct human evaluations on Yelp dataset with 100 randomly selected successful attacks produced by \ourmodel, CLARE, and MAYA. We evaluate these attacks in three aspects: (1) \textbf{Meaning preservation}: whether the attacks preserve the original meaning or not; (2) \textbf{Label preservation}: whether the modifications contradict the original sentiment or not; (3) We evaluate \textbf{fluency and grammaticality} via pairwise comparison: for each instance, we pair \ourmodel's attack with one by CLARE or MAYA. The human annotators are asked to either select the better one or rate them as equal. We average 6 responses for each sample. More details are in Appendix~\ref{app:human_evaluation_details}.

As shown in Table~\ref{tab:human_evaluation_results}, \ourmodel significantly outperforms CLARE and MAYA in terms of label consistency, i.e., 77.1\% vs. 49.8\%(CLARE) or 53.5\%(MAYA). This demonstrates the benefit of \setred{our whole architecture that utilizes both phrase-level infilling and sufficient constraints.} It's worth noting that all models struggle on preserving the textual meaning and less than 40\% of samples can \setred{retain} their original meaning. This is consistent with~\citealp{morris_reevaluating} in that semantic similarity metrics fail to maintain the actual meaning \setred{under most conditions}. On the fluency and grammaticality, \ourmodel is comparable to CLARE but is much better than MAYA (45\% vs. 26\%), since our context-aware blank-infilling is superior to paraphrasing each text piece independently. Finally, Table~\ref{tab:case1} compares adversarial attacks crafted by our model and other baselines. More case studies are provided in Appendix~\ref{app:additional_samples}.

\begin{table}[t!]
\setlength{\abovecaptionskip}{0.1cm}
\setlength{\belowcaptionskip}{-0.3cm}
\centering
\setlength{\tabcolsep}{1pt}
\small

\begin{tabular}{@{} l cccc @{}}
    \toprule
     & Yelp & AG News & MNLI & QNLLI
    \\\midrule
    1$^{st}$ & NP / 39.9\% & NP / 53.7\% & NP / 57.6\% & NP / 58.6\%\\
    2$^{nd}$ & ADJP / 17.5 \% & NNP / 27.0\%  & PP / 12.4\% & NNP / 16.4\%\\
    3$^{rd}$ &  VP / 16.6\% & PP / 9.5\% & NNP / 27.1\% & PP / 13.4\%\\
    \bottomrule
\end{tabular}

\caption{Top-3 phrase tags of attack phrases and their percentages on different datasets by \ourmodel.}
\label{tab:vulnerable_phrase}
\end{table}

\paragraph{Vulnerable phrase types.} We also analyze the three mostly attacked phrase types on each dataset. As shown in Table~\ref{tab:vulnerable_phrase}, noun phrases (NP) are the most vulnerable phrases over all datasets (more than 50\% on three datasets), while preposition phrases (PP) and proper noun phrases (NNP) are also commonly vulnerable.

\section{Analysis}
In this section, we conduct detailed analyses of \ourmodel, including ablation study (\secref{sec:ablation}), discussion of controllability in blank infilling (\secref{sec:controllable_ability}), and \setred{attacking} robust defense models (\secref{sec:robust_model}).

\subsection{Ablation Study}
\label{sec:ablation}

\begin{table}[t]
\setlength{\abovecaptionskip}{0.1cm}
\setlength{\belowcaptionskip}{-0.3cm}
\centering
\setlength{\tabcolsep}{1.5pt}
\small

\begin{tabular}{lrrrrr}
\toprule
\textbf{Module}  &
\textbf{ASR}$\uparrow$ & \textbf{DIS}$\downarrow$ & \textbf{BLEU}$\uparrow$ & \textbf{PPL}$\downarrow $ & \textbf{GER}$\downarrow$  \\
\midrule
\textsc{\ourmodel} & 98.4 & 0.17 & 0.78 & \textbf{56.8} & 0.33\\
\midrule
\textsc{\textit{w/o} phrase-level} & 97.7 & \textbf{0.09} & \textbf{0.84} & 92.8  & 0.68  \\
\textsc{\textit{w} all constituents}  & 98.2 & 0.16 & 0.79 & 58.1  & \textbf{0.29} \\
$\text{BERT}_{\text{base}}$ likelihood & 98.5 & 0.17 & 0.78 & \textbf{56.8} & 0.30 \\
\midrule
$\text{T5}_{\text{base}}$ infilling & 98.4 & 0.16 & 0.79 & 61.4 & 0.41 \\
$\text{GPT-2}_{\text{small}}$ infilling & \textbf{98.7} & 0.16 & 0.79 & 61.4 & 0.42 \\
\bottomrule
\end{tabular}

\caption{Results of the ablation study on Yelp dataset.}
\label{tab: abl}
\label{tab:abl}
\end{table}
We evaluate the effectiveness of each key component in \ourmodel based on the 1,000 random Yelp samples~\secref{sec:data}. We first study the phrase-level perturbation by replacing it with the word-level replacement used in Textfooler (\textsc{\textit{w/o} phrase-level}). As Table~\ref{tab: abl} shows, phrase-level perturbation has a larger attack search space which leads to better attack success rate, perplexity and grammar score. It also \setred{demonstrates} that attacking the constituents that are labeled as phrases is more effective than attacking all possible constituents. This is probably because phrases contain more critical and clear information to attack in classification tasks. 
In addition, we have not observed a significant difference between using BERT$_{\text{base}}$ and RoBERTa$_{\text{base}}$ for class-conditioned likelihood calculation, probably due to their similar architectures and shared knowledge. Finally, we comparing different blank-infilling methods using pre-trained $\text{BART}_{\text{base}}$ (\ourmodel), pre-trained $\text{T5}_{\text{base}}$~\cite{raffel_t5} and fine-tuned $\text{GPT-2}_{\text{small}}$ \cite{donahue_enabling}. Empirically,  $\text{BART}_{\text{base}}$ achieves the best overall performance. \setred{Moreover, we investigate the effectiveness of the label-preservation filter in \secref{sec:label_preservation_effc}}




\begin{figure}[t!]
    \setlength{\abovecaptionskip}{0.1cm}
    \setlength{\belowcaptionskip}{-0.4cm}
    \centering
    \includegraphics[width=0.8\linewidth]{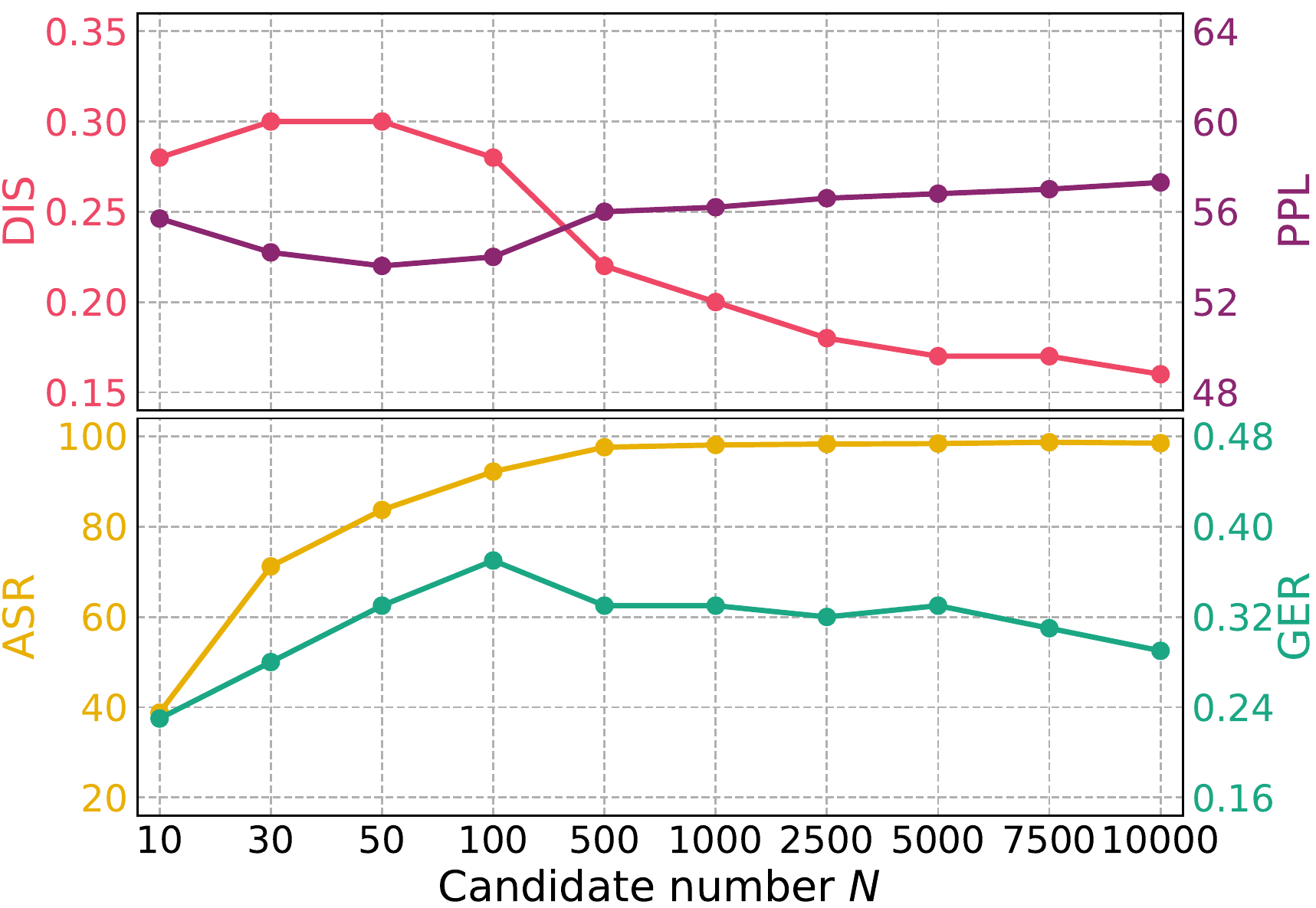}
    \caption{\textbf{ASR}, \textbf{GER}, \textbf{DIS} and \textbf{PPL} results by controlling different candidates numbers $N$ in \ourmodel.}
    \label{fig:candidate_number}
\end{figure}

\begin{figure}[t!]
    \setlength{\abovecaptionskip}{0.1cm}
    \setlength{\belowcaptionskip}{-0.4cm}
    \centering
    \includegraphics[width=0.8\linewidth]{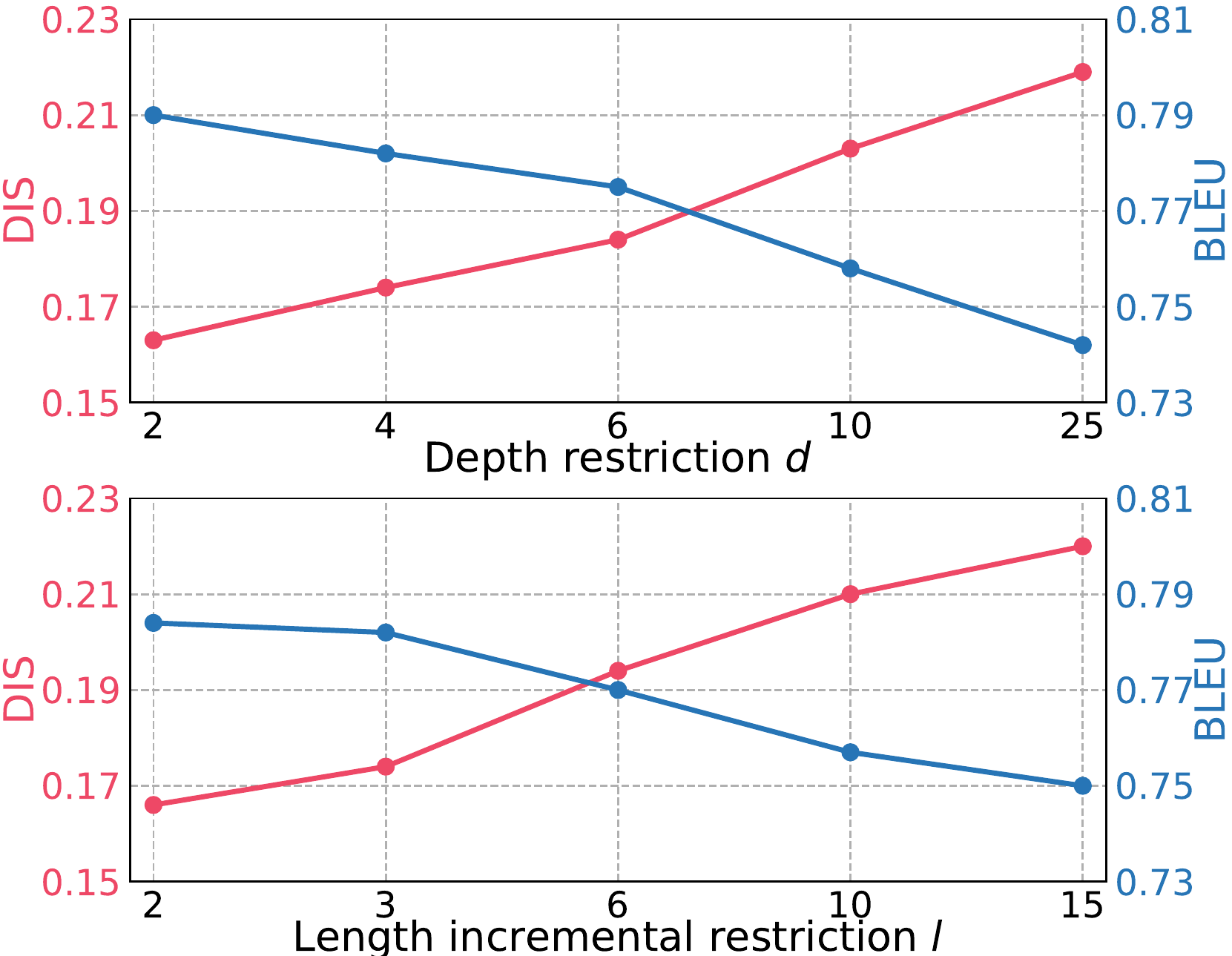}
    \caption{\textbf{DIS} and \textbf{BLEU} results by varying depth restrictions $d$ (upper) and length increments $l$ (bottom).}
    \label{fig:effect_depth_length}
\end{figure}

\subsection{Sensitivity and Controllability Analysis}
\label{sec:controllable_ability}

In this section, we investigate how the outputs can be controlled by hyper-parameters in \ourmodel on Yelp dataset. We first study how the generated samples can be impacted by varying the candidate number $N$ in blank-infilling. As shown in Figure~\ref{fig:candidate_number},
the success rate (\textbf{ASR}) increases with the increase of $N$ but starts to saturate when $N\geq500$, while the grammar errors (\textbf{GER}) stay quite consistent. Meanwhile, the edit distance (\textbf{DIS}) drops significantly but the perplexity (\textbf{PPL}) only increases slightly when $N$ increases. Based on these observations, we choose $N=5000$ in our experiments for the best trade-off among these aspects. 

In addition, we explore how the syntactic tree depth $d$ in target phrases and the incremental length $l$ for perturbed phrases can affect the modification degree. In Figure~\ref{fig:effect_depth_length}, with the increase of $d$ or $l$, more modifications are introduced and the textual similarity decreases, which undermines the validity of adversarial samples, as larger $d$ and $l$ \setred{may} augment the attacking space with longer but unnecessary perturbations. Hence, we choose relatively small $d$ and $l$ to ensure the attack is more controllable. Note that both $d$ and $l$ exhibit a slight impact on the success rate \setred{(Refer to Appendix~\ref{app:additional_results} for detailed results).}





\begin{table}[t]
\setlength{\abovecaptionskip}{0.1cm}
\setlength{\belowcaptionskip}{-0.2cm}
\centering
\setlength{\tabcolsep}{4pt}
\small
\begin{tabular}{@{} l ccccc @{}}
\toprule
Method & \textbf{ASR}$\uparrow$ & \textbf{DIS}$\downarrow$ & \textbf{BLEU}$\uparrow$ & \textbf{PPL}$\downarrow$ & \textbf{GER}$\downarrow$ \\
\midrule
TextFooler & 64.2 & 0.23 & 0.59 & 185.5 & 1.39 \\
CLARE & 92.5 & \textbf{0.12} & \textbf{0.81} & 76.1 & \textbf{0.15} \\
\arrayrulecolor{black!50}\specialrule{.3pt}{1pt}{1pt}
MAYA & 98.6 & 0.57 & 0.33 & 82.1 & 3.51 \\
\ourmodel & \textbf{99.2} & 0.28 & 0.67 & \textbf{55.7} & 0.39 \\
\bottomrule
\end{tabular}

\caption{Results of \ourmodel and baselines attacking the robust defense model TAVAT on Yelp dataset. }
\label{tab:robust}
\end{table}

\subsection{Attacking Robust Defense Model}
\label{sec:robust_model}
In this section, we examine whether existing robust defense models can defend \ourmodel attack which is beyond word-level perturbations. We apply a robust $\text{BERT}_{\text{base}}$ defense model trained via TAVAT~\cite{li2021tavat} to \setred{defend} the attacks from \ourmodel and baseline models, which is designed for word-level attacks. Comparing Table~\ref{tab:robust} with Table~\ref{tab:main_results}, both the edit distance and BLEU get worse when attacking the defense model, showing that the defense model is harder to attack. Meanwhile, two word-level attacks have a significant attack success rate drop, e.g., 94.5\% to 64.2\% on TextFooler. On the contrary, \ourmodel still can achieve the best 99.2\% attack success rate with sufficient textual similarity and grammar errors, outperforming MAYA \setred{in} every aspect. This suggests that \ourmodel raises a new robustness issue on current defense models.

\section{Related Work}
\paragraph{Textual Adversarial Attack}
Growing interest is devoted to generating textual adversarial samples via perturbation at various levels.
Some early works use misspelling tokens in character-level~\cite{liang_character,ebrahimi_character_flip,li_textbugger}, but they can be easily defended by spell checking tools~\cite{pruthi_combat_misspelling,zhou_discriminate,jones_robust_encoding}.
Recent mainstream of studies try to misguide models via word-level perturbations, e.g., synonym/semantic neighbor substitution~\cite{alzantot_generating,jin_textfooler,ren_pwws,zhang_generating}, replacement by \setred{pre-trained} masked language models~\cite{li_bert_attack,zhang_generating}, or combing \setred{replacement with} operations like \setred{insertion and merging}~\cite{li_clare}. These methods usually attempt to preserve the semantic similarity for better fluency and grammaticality, but their perturbations are limited to independent single words.

Sentence-level attacks have also been studied to generate new texts via paraphrasing or GAN-based generation~\cite{iyyer_paraphrase,wang_cat_gan,zou_reinforced,wang_t3, qi_style}, but their drastic modifications on the text structure make it harder to maintain a satisfactory textual quality.
Very recently, phrase-level perturbations are considered in evaluating the performance of syntactic parsing~\cite{zheng_phrase_parser}, or involved in a multi-granularity textual attack model MAYA~\cite{chen_maya}. 
Unlike MAYA, \ourmodel only focuses on unified phrase-level attacks without multiple sub-modules, which require simpler and fewer modifications while benefiting better performance.

\paragraph{Blank Infilling}
Large-scale pre-trained language models such as BERT~\cite{devlin_bert} and RoBERTa~\cite{liu_roberta} have shown their capability of filling masked single tokens~\cite{wang_bert_generate,ghazvininejad_mask} but they cannot handle variable-length masks. Although autoregressive generative models such as GPT~\cite{radford_gpt} or GPT2~\cite{radford_gpt2} can produce \setred{outputs} with arbitrary \setred{lengths}, they only condition information from a single direction \setred{and ignore some surrounding texts}. To enable GPT models to fill in blanks, \citeauthor{donahue_enabling} proposed to fine-tune them with sequences concatenating manually-masked texts and missing texts. Recently, autoencoder-decoder models such as T5~\cite{raffel_t5} and BART~\cite{lewis_bart}, \setred{which are} trained using \setred{text} infilling losses make it possible to fill the blanks within the context in a more flexible form~\cite{shen_blank_language_models}. 

\section{Conclusion}
We present a new phrase-level textual adversarial attack, \ourmodel, which produces richer and higher-quality phrase-level perturbations than the widely studied word-level attacks.
It utilizes contextualized blank-infilling to generate perturbations by a pre-trained language model and thus well preserves the \setred{fluency, and grammaticality}. We additionally develop a label-preservation filter \setred{based on the likelihood given by class-conditioned language models, trying to }to keep the ground-truth labels intact. Extensive experiments show the effectiveness of \ourmodel and its advantages over baselines on different NLP tasks.

\bibliography{anthology,custom}
\bibliographystyle{acl_natbib}

\clearpage
\newpage
\appendix

\section{Implementation Details}

\subsection{Details of \ourmodel}

\paragraph{Basic infrastructure}


We use PyTorch as the backbone of our implementation, along with Huggingface-Transformers\footnote{\url{https://github.com/huggingface/transformers}} for the implementation of victim models and likelihood estimation models, while Fairseq for the implementation of blank-infilling model BART\footnote{\url{https://github.com/pytorch/fairseq}}. We list the hyperparameters used in our model in Table~\ref{tab:hyparameters}, all of them are determined empirically based on both attack success rate and textual quality. It takes \textbf{about 160 minutes to generate 100 adversarial samples} on Yelp dataset using \ourmodel on a single NVIDIA GTX 1080 Ti GPU.

\paragraph{Select phrase candidates.} 
We use the parser from Stanford CoreNLP\footnote{\url{https://stanfordnlp.github.io/CoreNLP/}} toolkit for the syntactic tree parsing. We consider parsed nodes with the syntactic tags in Table~\ref{tab:syntactic_tags} as the possible root node of a phrase.




\begin{table}[htbp]
\setlength{\belowcaptionskip}{-0.2cm}
\centering
\small
\begin{tabularx}{\linewidth}{rl}
\toprule
Depth restriction of phrase syntactic tree $d$ & 4\\
Length incremental restriction for substitutions $l$ & 3\\
Maximum perturbation number T &  11 \\
Likelihood ratio filter threshold $\delta$ & 1 \\
Substitution candidates number $N$ & 5000 \\
\bottomrule
\end{tabularx}
\caption{All hyperparameters used in \ourmodel.}
\label{tab:hyparameters}
\end{table}

\begin{table}[htbp]
\setlength{\belowcaptionskip}{-0.2cm}
\centering
\small
\begin{tabularx}{\linewidth}{lX}
\toprule
Tags & ADJP, ADVP, CONJP, NP, NNP, PP, QP, VP, WHADJP, WHADVP, WHNP, WHVP \\
\bottomrule
\end{tabularx}
\caption{Syntactic tags that will be selected as the root of a phrase.}
\label{tab:syntactic_tags}
\end{table}

\paragraph{Blank filling with language model.}
In our default setting, we \setred{directly apply} the original BART$_{\rm base}$ model\footnote{\url{https://dl.fbaipublicfiles.com/fairseq/models/bart.base.tar.gz}} with 6 encoder and decoder layers and 140M parameters. During infilling, the target phrase $\mathbf{a}$ will be replaced with a special symbol ``<mask>'', then the model will fill this blank with variable-length context. The Top-K sampling strategy is used during generation, where we set $k=50$ and repeat this procedure several times to collect enough phrase substitution candidates. Since BART$_{\rm base}$ implement blank filling via reconstructing the whole sentence, where the text excluding the blank part after filling may not be the same as the original one,  we simply only reserve the reconstructed sentences which keep the text excluding the blank part unchanged. 


In the variations with other blank-infilling models, we use GPT-2$_{\rm small}$ model\footnote{\url{https://huggingface.co/gpt2}} with 124M parameters or T5$_{\rm base}$ model\footnote{\url{https://huggingface.co/t5-base}} with 220M parameters, both have 12 layers. Since the original GPT-2$_\text{small}$ model is not suitable for blank infilling, we enable GPT-2 model to implement this task by finetuing on Yelp training corpus using method proposed by \citeauthor{donahue_enabling}, running the code provide by the authors\footnote{\url{https://github.com/chrisdonahue/ilm}}.

\begin{table}[htbp]
\setlength{\abovecaptionskip}{0.1cm}
\setlength{\belowcaptionskip}{-0.2cm}
\centering
\setlength{\tabcolsep}{4.0pt}
\small{
\begin{tabular}{@{} lccrrr @{}}
\toprule

\textbf{Dataset}  &
\textbf{PPL before} & \textbf{PPL after} 
\\\midrule

Yelp  &  11.44     & 6.29\\
AG News &  9.33  & 3.64 \\
\midrule

MNLI &  6.16 & 3.74\\
QNLI & 5.14  & 5.00\\
\bottomrule
\end{tabular}
}
\caption{Perplexities of our fine-tuned masked language models for likelihood estimation, before and after fine-tuning on the validation set of each dataset (prepending our predefined special label token).}
\label{tab:mlm_results}
\end{table}

\paragraph{Models for calculating likelihood.}
We apply RoBERTa$_{\rm base}$\footnote{\url{https://huggingface.co/roberta-base}} model fine-tuned on the corresponding training set of each attack dataset in this stage, which has 12 layers and 125M parameter. Then we fine-tune the label-conditioned masked language models on different datasets as follows to make them better fit the specific domain.
\begin{itemize}[wide=0\parindent, noitemsep, topsep=0pt]
\item Classification datasets (Yelp, AG News): Since each sample is usually long, we split a sample into several short sentences as the input for fine-tuning. To avoid the conditions that some short sentences may be irrelevant or contradict the overall label $y$, we employ a classifier to make predictions on these sentences and only remain sentences with the confidence value of $y$ higher than 0.99. Such a sentence along with the special label token ``\textit{<Label>}'' corresponding to the overall label $y$ will form a new sample for fine-tuning, whose input format is ``\textit{<Label>} \ \textit{Sentence}''.
\item NLI datasets (MNLI, QNLI): samples in these datasets are usually a pair of short sentences, so we will not split them. The input format for these datasets is ``\textit{<Label>} \ \textit{SentenceA}  \ \textit{</s>} \ \textit{</s>} \ \textit{SentenceB}'', where ``\textit{</s></s>}'' is the separation symbol in RoBERTa. 
\end{itemize}
Then we will randomly mask some tokens in these samples to fine-tune a masked language model conditioned on labels, the batch size is 8, and the learning rate is $5e^{-5}$ with AdamW optimizer. The PPL before and after fine-tuning is shown in Table~\ref{tab:mlm_results}, demonstrating the effectiveness of this procedure.

When predicting the likelihood of a perturbation, we will concatenate the label of the original sample with the masked perturbed sequence as the input, similar to samples in fine-tuning. When attacking Yelp and AG News datasets, we also only use the sub-sentence containing the perturbation, rather than the whole text.

\paragraph{Metrics.} To obtain the edit distance (DIS) metric, we utilize the open-source tool\footnote{\url{https://github.com/roy-ht/editdistance}} to calculate it in token-level, i.e. how many words need to be edited to transform a text into another one and then normalized by the text length. In addition, we employ Toolkit in NLTK\footnote{\url{https://www.nltk.org/_modules/nltk/translate/bleu_score.html}} to calculate BLEU metrics between adversarial samples and the corresponding original samples.

\subsection{Details of Victim Models}

\paragraph{BERT models.}
All BERT victim models are based on BERT$_{\text{base}}$\footnote{\url{https://huggingface.co/bert-base-uncased}}, which contains 110M parameters with 12 layers. A linear layer is added for classification, which takes the representation of ``[CLS]'' token in the head of a sequence as the input. We then fine-tune victim models on each dataset using batch size 32 and the learning rate $2e^{-5}$ for 3 epochs. The final model after 3 epochs will be saved and used as the victim model $F$ on each dataset.

\paragraph{Train robust models using TAVAT.}

The robust models are also based on BERT$_{\text{base}}$ with a linear layer added for classification. We fine-tune the model using an adversarial training method TAVAT proposed by \citeauthor{li2021tavat} which is a token-level gradient accumulation of perturbations, by running code provided by the authors\footnote{\url{https://github.com/LinyangLee/Token-Aware-VAT}} with all default hyper-parameters. During finetuning, perturbations guided by gradient are applied to the embedding space and models are trained using these perturbed data.

\subsection{Details of baseline MAYA}
MAYA has three variants: MAYA, MAYA$_\pi$ and MAYA$_\text{bt}$. We select MAYA as our baseline since overall it obtains the best attack success rate and perplexity performance.

\subsection{Possible Limitations of Our Model}

The label-preservation filter in our \ourmodel model utilizes label-conditioned masked language models, which need to be fine-tuned on a labeled corpus with sufficient data. Therefore, the performance of \ourmodel may drop on datasets that have limited number of labeled samples. In addition, it takes about 1 minute for our model to generate one adversarial sample using BERT as the victim model, so \ourmodel is not applicable for conditions with low computational resources.

\section{Additional Qualitative Samples}\label{app:additional_samples}

We introduce some additional adversarial samples generated by our \ourmodel model, along with three baselines, TextFooler, CLARE, MAYA, on four datasets, Yelp, AG News, MNLI, QNLI, in Table~\ref{tab:addition_case1}, Table~\ref{tab:addition_case2}, and Table~\ref{tab:addition_case3}.

\begin{table*}[htbp]
\small
\setlength{\abovecaptionskip}{0.1cm}
\setlength{\belowcaptionskip}{-0.2cm}
\centering
\begin{tabularx}{\textwidth}{p{3cm}X}
\toprule
\textbf{Yelp (Positive)} &  \ Excellent food at this out of the way place. Portions very large and fresh. I want to try everything on the menu. Plan to go back every weekend until I've tried all menu items. Coffee was also delicious and friendly servers\\
\noalign{\vskip 2pt}\hdashline\noalign{\vskip 2pt}
\textbf{TextFooler (Negative)} & \ \setblue{Outstanding foods} at this out of the way place. Portions very large and \setblue{mild}. I want to \setblue{dabbled whatsoever} on the menu. Plan to go back \setblue{all monday} until I've \setblue{attempted} all menu items. Coffee was also \setblue{peachy} and friendly servers\\
\noalign{\vskip 2pt}\hdashline\noalign{\vskip 2pt}
\textbf{CLARE (Negative)} & \ \setgreen{Incredible} food at this out of \setgreen{control} place. Portions \setgreen{plentiful and plentiful}. I want to try \setgreen{something} on the menu. Plan to go back \setgreen{mid} weekend until I've tried all menu items. Coffee was \setgreen{fairly comforting} and friendly servers \\
\noalign{\vskip 2pt}\hdashline\noalign{\vskip 2pt}
\textbf{MAYA (Negative)} & \ \setredd{The} food at this out of the way place. Portions very large and \setredd{expensive}. I want to try everything on the menu. Plan to go back every weekend until I've tried all menu items. Coffee was also \setredd{cheap} and friendly servers \\
\noalign{\vskip 2pt}\hdashline\noalign{\vskip 2pt}
\textbf{\ourmodel (Negative)} &  \ \setorange{I had nothing but fun} at this out of the way place. Portions very large and fresh. I want to try everything on the menu. Plan to go back every weekend until I've tried all menu items. Coffee was also delicious and friendly servers\\

\toprule

\textbf{Yelp (Positive)} &  \ I love this place. I love everything there except the kabsa rice but that's just me. Burgers are good. They pile on the veggies. Owner is nice. Freshly made food always has my mouth watering .\\
\noalign{\vskip 2pt}\hdashline\noalign{\vskip 2pt}
\textbf{TextFooler (Negative)} &  \ I \setblue{aime} this place. I love everything there except the kabsa rice but that's just me. Burgers are good. They pile on the veggies. Owner is nice. Freshly made food always has my mouth watering.\\
\noalign{\vskip 2pt}\hdashline\noalign{\vskip 2pt}
\textbf{CLARE (Negative)} & \ I \setgreen{hate} this place. I love everything there except the kabsa rice but that's just me. Burgers are good. They pile on the veggies. Owner is nice. Freshly made food always has my mouth watering. \\
\noalign{\vskip 2pt}\hdashline\noalign{\vskip 2pt}
\textbf{MAYA (Negative)} & \ I \setredd{know} this place. I love everything there except the kabsa rice but that's just me. Burgers are good. They pile on the veggies. Owner is nice. Freshly made food always has my mouth watering. \\
\noalign{\vskip 2pt}\hdashline\noalign{\vskip 2pt}
\textbf{\ourmodel (Negative)} & \ \setorange{I can't recommend Aptopia enough}. I love everything there except the kabsa rice but that's just me. Burgers are good. They pile on the veggies. Owner is nice. Freshly made food always has my mouth watering.\\
\bottomrule
\end{tabularx}

\begin{tabularx}{\textwidth}{p{3cm}X}
\toprule
\textbf{AG News (Sci-tech)} & \ Scientists Discover Ganymede has a Lumpy Interior. Jet Propulsion Lab--Scientists have discovered irregular lumps beneath the icy surface of Jupiter's largest moon, Ganymede. These irregular masses may be rock formations, supported by Ganymede's icy shell for billions of years...\\
\noalign{\vskip 2pt}\hdashline\noalign{\vskip 2pt}
\textbf{TextFooler (World)} & \ \setblue{Researchers Unmask Deimos} has a Lumpy \setblue{Indoors}. Jet \setblue{Rotor Laboratories}--\setblue{Searchers} have discovered irregular \setblue{clods into} the icy surface of Juniper's largest moon, \setblue{Jupiter}. These irregular masses maggio be rock formations, \setblue{contributions} by \setblue{Enceladus}'s icy shell for billions of years...\\
\noalign{\vskip 2pt}\hdashline\noalign{\vskip 2pt}
\textbf{CLARE (Business)} &  \ Scientists \setgreen{Know} Ganymede has a Lumpy Interior. \setgreen{Credit} Jet Propulsion Lab--\setgreen{Featured} Scientists have discovered irregular lumps beneath the icy surface of Jupiter's largest moon, Ganymede. These irregular masses may be rock formations, supported by Ganymede's icy shell for billions of years... \\
\noalign{\vskip 2pt}\hdashline\noalign{\vskip 2pt}
\textbf{MAYA (World)} & \ Scientists Discover Ganymed has a Lumpy Interior. \setredd{Scientists have discovered irregular lumps under the icy surface of jupiter's largest moon..} These irregular masses may be rock formations, supported by ganymede's icy shell for billions of years... \\
\noalign{\vskip 2pt}\hdashline\noalign{\vskip 2pt}
\textbf{\ourmodel (World)} & \ Scientists Discover Ganymede has a Lumpy Interior. \setorange{JPL-Caltech STOCKHOLM}--Scientists have discovered irregular lumps beneath the icy surface of Jupiter's largest moon, Ganymede. These irregular masses may be rock formations, supported by Ganymede's icy shell for billions of years...\\
\toprule

\textbf{AG News (Sport)} & \ Giddy Phelps Touches Gold for First Time. Michael Phelps won the gold medal in the 400 individual medley and set a world record in a time of 4 minutes 8. 26 seconds.\\
\noalign{\vskip 2pt}\hdashline\noalign{\vskip 2pt}
\textbf{Textfooler (World)} & \ \setblue{Dazzled} Phelps \setblue{Hits} Gold for \setblue{Premiere} Time. Michael Phelps won the gold \setblue{trophy} in the 400 \setblue{personal} medley and set a world record in a \setblue{hours} of 4 \setblue{record} 8. 26 seconds.\\
\noalign{\vskip 2pt}\hdashline\noalign{\vskip 2pt}
\textbf{CLARE (World)} &  \ Giddy Phelps Touches Gold for First Time. Michael Phelps won the gold medal in the 400 individual medley and set a world record in a time of 4 minutes 8 \setgreen{...} \\
\noalign{\vskip 2pt}\hdashline\noalign{\vskip 2pt}
\textbf{MAYA (World)} & \ Giddy Phelps Touches Gold for First Time. Michael Phelps \setredd{the gold medal in the 400 individual medley was won by him in a world record time of 4 minutes 8 seconds..} \\
\noalign{\vskip 2pt}\hdashline\noalign{\vskip 2pt}
\textbf{\ourmodel (World)} & \ \setorange{Swimmers: Phelps} Touches Gold for First Time. Michael Phelps won the gold medal in the 400 individual medley and set a world record in a time of 4 minutes 8.26 seconds.\\
\bottomrule
\end{tabularx}

\caption{\label{tab:addition_case1} Adversarial examples generated by different models on Yelp and AG News dataset, perturbations are colored.}
\end{table*}

\begin{table*}[htbp]
\small
\setlength{\abovecaptionskip}{0.1cm}
\setlength{\belowcaptionskip}{-0.2cm}
\centering

\begin{tabularx}{\textwidth}{p{3cm}X}
\toprule
\textbf{MNLI} \newline \textbf{(Neutral)} &  
\textbf{Premise} The last politician to propose making driving more expensive was Al Gore, who fought to include a small energy tax--which would have included gasoline--in the Clinton administration's 1993 economic plan.
\newline
\textbf{Hypothesis} Al Gore is still making proposals for making driving more expensive. \\
\noalign{\vskip 2pt}\hdashline\noalign{\vskip 2pt}
\textbf{TextFooler} \newline \textbf{(Contradiction)} & 
\textbf{Premise} The last \setblue{policies} to propose making driving more expensive was Al Gore, who fought to include a small energy tax--which would have included gasoline--in the Clinton administration's 1993 economic plan.
\newline
\textbf{Hypothesis} Al Gore is still making proposals for making driving more expensive. \\
\noalign{\vskip 2pt}\hdashline\noalign{\vskip 2pt}
\textbf{CLARE} \newline \textbf{(Contradiction)} &  
\textbf{Premise} The last politician to propose making driving more expensive was Al Gore, who \setgreen{moved} to include a small energy tax--which would have included gasoline--in the Clinton administration's 1993 economic plan. 
\newline
\textbf{Hypothesis} Al Gore is still making proposals for making driving more expensive. \\
\noalign{\vskip 2pt}\hdashline\noalign{\vskip 2pt}
\textbf{MAYA} \newline \textbf{(Contradiction)} & 
\textbf{Premise} The last politician \setredd{propose to make driving more expensive} was AI Gore, who fought to include a small energy tax--which would have included gasoline--in the clinton administration's 1993 economic plan.
\newline
\textbf{Hypothesis} Al Gore is still making proposals for making driving more expensive. \\
\noalign{\vskip 2pt}\hdashline\noalign{\vskip 2pt}
\textbf{\ourmodel} \newline \textbf{(Contradiction)} & 
\textbf{Premise} The last politician to propose making driving more expensive was \setorange{his predecessor Senator Al Gore}, who fought to include a small energy tax--which would have included gasoline--in the Clinton administration's 1993 economic plan.
\newline
\textbf{Hypothesis} Al Gore is still making proposals for making driving more expensive. \\

\toprule

\textbf{MNLI} \newline \textbf{(Entailment)} &  
\textbf{Premise} So uh but but uh it runs fine all you have it's just very thirsty if I just keep the oil in it seems to be okay but you know that's a sign that I'm going to have to do something sooner or later
\newline
\textbf{Hypothesis} It runs well, but I think I might have to do some work on it. \\
\noalign{\vskip 2pt}\hdashline\noalign{\vskip 2pt}
\textbf{TextFooler} \newline \textbf{(Neutral)} & 
\textbf{Premise} So uh but but uh it runs fine all you have it's just very thirsty if I just keep the \setblue{petroleum} in it seems to be okay but you know that's a sign that I'm going to have to do \setblue{nothings shortly} or later
\newline
\textbf{Hypothesis} It runs well, but I think I might have to do some work on it. \\
\noalign{\vskip 2pt}\hdashline\noalign{\vskip 2pt}
\textbf{CLARE} \newline \textbf{(Neutral)} &  
\textbf{Premise} So uh but but uh it runs fine all you have it's just very thirsty if I just \setgreen{drink} the oil in it seems to be okay but you know that's a sign that I'm going to have to do \setgreen{nothing} sooner or later 
\newline
\textbf{Hypothesis} It runs well, but I think I might have to do some work on it. \\
\noalign{\vskip 2pt}\hdashline\noalign{\vskip 2pt}
\textbf{MAYA} \newline \textbf{(Neutral)} & 
\textbf{Premise} So uh but but uh it runs fine all you have it's just very thirsty \setredd{if it seems to be okay , but i'm going to have to do something soon or later.} 
\newline
\textbf{Hypothesis} It runs well, but I think I might have to do some work on it. \\
\noalign{\vskip 2pt}\hdashline\noalign{\vskip 2pt}
\textbf{\ourmodel} \newline \textbf{(Neutral)} & 
\textbf{Premise} So uh but but uh it runs fine all you have it's just very thirsty if I just keep \setorange{it that's all I got} in it seems to be okay but you know that's a sign that I'm going to have to do something sooner or later
\newline
\textbf{Hypothesis} It runs well, but I think I might have to do some work on it. \\
\bottomrule
\end{tabularx}

\caption{\label{tab:addition_case2} Adversarial examples generated by different models on MNLI dataset, perturbations are colored.}
\end{table*}

\begin{table*}[htbp]
\small
\setlength{\abovecaptionskip}{0.1cm}
\setlength{\belowcaptionskip}{-0.2cm}
\centering

\begin{tabularx}{\textwidth}{p{3cm}X}
\toprule
\textbf{QNLI} \newline \textbf{(Entailment)} & \textbf{Premise} What are some of the sets or ideals most school systems follow? 
\newline
\textbf{Hypothesis} Such choices include curriculum, organizational models, design of the physical learning spaces (e.g. classrooms), student-teacher interactions, methods of assessment, class size, educational activities, and more.\\
\noalign{\vskip 2pt}\hdashline\noalign{\vskip 2pt}
\textbf{TextFooler} \newline \textbf{(Not Entailment)} & \textbf{Premise} What are some of the sets or ideals most school systems follow? 
\newline
\textbf{Hypothesis} Such choices include curriculum, \setblue{organizes storyboards}, design of the \setblue{tangible} learning spaces (e.g. classrooms), student-teacher interactions, methods of assessment, class size, educational activities, and more.\\
\noalign{\vskip 2pt}\hdashline\noalign{\vskip 2pt}
\textbf{CLARE} \newline \textbf{(Not Entailment)} &  \textbf{Premise} What are some of the sets or ideals most school systems follow? 
\newline
\textbf{Hypothesis} Such choices \setgreen{affect} curriculum, organizational models, design of the physical learning spaces (e.g. classrooms), student-teacher interactions, methods of assessment, class size, educational activities, and more. \\
\noalign{\vskip 2pt}\hdashline\noalign{\vskip 2pt}
\textbf{MAYA} \newline \textbf{(Not Entailment)} & \textbf{Premise} What are some of the sets or ideals most school systems follow? 
\newline
\textbf{Hypothesis} Such choices \setredd{the design of the physical learning spaces should include curriculum, organizational models, and methods of assessment.., and class size.., and educational activities..} \\
\noalign{\vskip 2pt}\hdashline\noalign{\vskip 2pt}
\textbf{\ourmodel} \newline \textbf{(Not Entailment)} & \textbf{Premise} What are some of the sets or ideals most school systems follow? 
\newline
\textbf{Hypothesis} Such choices include curriculum, \setorange{use of a teacher's manual}, design of the physical learning spaces (e.g. classrooms), student-teacher interactions, methods of assessment, class size, educational activities, and more.\\

\toprule

\textbf{QNLI} 
\newline
\textbf{(Entailment)} & \textbf{Premise} What to the migrating birds usually follow? 
\newline
\textbf{Hypothesis} These routes typically follow mountain ranges or coastlines, sometimes rivers, and may take advantage of updrafts and other wind patterns or avoid geographical barriers such as large stretches of open water.\\
\noalign{\vskip 2pt}\hdashline\noalign{\vskip 2pt}
\textbf{Textfooler} \newline \textbf{(Not Entailment)} & \textbf{Premise} What to the migrating birds usually follow? 
\newline
\textbf{Hypothesis} These routes \setblue{seldom} follow \setblue{colina telemetry} or coastlines, sometimes rivers, and may take advantage of updrafts and other wind \setblue{diagrams} or avoid \setblue{spatial} \setblue{separating} such as large stretches of \setblue{commencement water}.\\
\noalign{\vskip 2pt}\hdashline\noalign{\vskip 2pt}
\textbf{CLARE} \newline \textbf{(Not Entailment)} &  \textbf{Premise} What to the migrating birds usually follow? 
\newline
\textbf{Hypothesis} These routes \setgreen{cannot} follow \setgreen{continental} ranges or coastlines, \setgreen{connect rivers}, and may take advantage of updrafts and other wind patterns or avoid geographical barriers such as large stretches of open water. \\
\noalign{\vskip 2pt}\hdashline\noalign{\vskip 2pt}
\textbf{MAYA} \newline \textbf{(Not Entailment)} & \textbf{Premise} What to the migrating birds usually follow? 
\newline
\textbf{Hypothesis} These \setredd{lines} typically \setredd{connect} mountain ranges or coastlines, sometimes rivers, and may take advantage of updrafts and other wind patterns or avoid geographical barriers such as large stretches of open water. \\
\noalign{\vskip 2pt}\hdashline\noalign{\vskip 2pt}
\textbf{\ourmodel} \newline \textbf{(Not Entailment)} & \textbf{Premise} What to the migrating birds usually follow ? 
\newline
\textbf{Hypothesis} These routes typically \setorange{take advantages from} larger mountain ranges or coastlines, sometimes rivers, and may take advantage of updrafts and other wind patterns or avoid geographical barriers such as large stretches of open water.\\
\bottomrule
\end{tabularx}

\caption{\label{tab:addition_case3} Adversarial examples generated by different models on QNLI dataset, perturbations are colored.}
\end{table*}

\newpage
\section{Additional Results}
\label{app:additional_results}

\subsection{Effectiveness of Label-Preservation Filter}
\label{sec:label_preservation_effc}
\setred{Since human evaluation on ablation study of the label-preservation filter is costly, to verify its effect, we turn to use adversarial samples derived from the attack on the Yelp dataset.
Here, we train models with either only adversarial examples or a combination of original and adversarial examples to evaluate the label correctness of adversarial samples.
Since the labels of the Yelp test set are carefully annotated, 
the accuracy on the Yelp test set of each trained classifier can be regarded as a metric to reflect the label-preservation capability of the crafted adversarial examples. A better label-preservation of adversarial samples will introduce less noise during training and thus results in preferable performance on the test set. According to the results shown in Table~\ref{tab:effect_label_preservation}, models trained using samples from our complete \ourmodel model with the label preservation filter shows higher accuracy than ones trained on samples derived from ablated \ourmodel without such a filter, especially when only training the classifiers based on adversarial samples. Therefore, it demonstrates that using the label-preservation filter can obtain adversarial examples with better label consistency.}      

\begin{table}[htbp]
\setlength{\tabcolsep}{1mm}
\centering
\begin{tabular}{@{} l cc @{}}
\toprule
 \textbf{Module} & \textbf{Only Adv} & \textbf{Combined}  \\
\midrule
\ourmodel & 88.8\% & 89.2\% \\
\quad \textit{w/o label-preservation} & 87.3\% & 88.6\%  \\
\bottomrule
\end{tabular}
\caption{Accuracy on the Yelp test set of models that are trained on barely adversarial examples (\textbf{Only Adv}) or a combination of original and adversarial examples (\textbf{Combined}). The adversarial samples come from our complete \ourmodel model, or  its ablation without label-preservation filter.  }
\label{tab:effect_label_preservation}
\end{table}

\subsection{Additional Results on Controllable Ability}
We list the full results of \secref{sec:controllable_ability} about controllable ability on Yelp dataset in Table~\ref{tab:effect_candidate_num}, Table~\ref{tab:effect_depth}, and Table~\ref{tab:effect_length}, for the effects of candidate number $N$ during infilling, syntactic tree depth restriction for phrase candidates $d$ and length incremental restriction for substitution $l$ on our \ourmodel model, respectively. It can be found that all $N$, $d$, and $l$ can control the performance of \ourmodel in different aspects. Surprisingly, the grammar error decreases while $d$ is increasing. We attribute this to the fact that the modification range of the phrase is extended as $d$ increases, such that the infilling text is less likely to be essentially conditioned on the surrounding context and fewer grammar errors occur at the boards between blanks and rest texts.

\begin{figure}[!h]
\setlength{\belowcaptionskip}{-0.2cm}
    \centering
    \includegraphics[width=\linewidth]{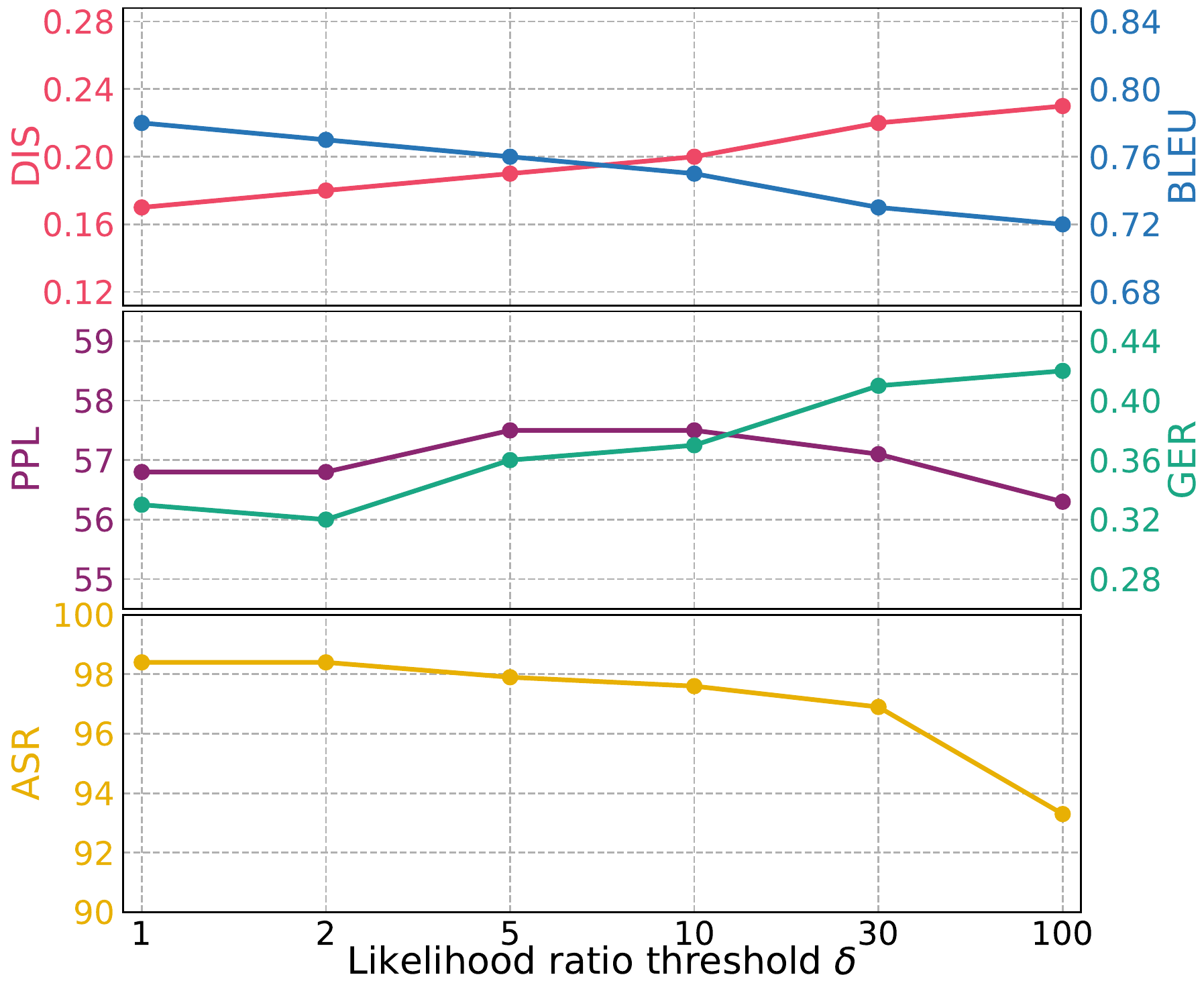}
    \caption{The performance of \ourmodel on Yelp dataset using different likelihood ratio threshold in label-preservation filter, in terms of all 5 metrics.}
    \label{fig:effect_likelihood}
\end{figure}

We also test the effects of different likelihood ratio threshold $\delta$ on Yelp dataset, which is illustrated in Figure~\ref{fig:effect_likelihood}. A larger threshold $\delta$ may result in a lower attack success rate, more modifications, and a worse textual quality. 

\begin{table}[htbp]
\setlength{\tabcolsep}{4pt}
\centering
\begin{tabular}{lccccc}
\toprule
$N$ & \textbf{ASR}$\uparrow$ & \textbf{DIS}$\downarrow$ & \textbf{BLEU}$\uparrow$ & \textbf{PPL}$\downarrow$ & \textbf{GER}$\downarrow$ \\
\midrule
10 & 38.8 & 0.28 & 0.66 & 55.7 & 0.23 \\
30 & 71.2 & 0.30 & 0.63 & 54.2 & 0.28 \\
50 & 83.7 & 0.30 & 0.63 & 53.6 & 0.33 \\
100 & 92.2 & 0.28 & 0.66 & 54.0 & 0.37 \\
500 & 97.6 & 0.22 & 0.73 & 56.0 & 0.33 \\
1,000 & 98.1 & 0.20 & 0.75 & 56.2 & 0.33 \\
2,500 & 98.3 & 0.18 & 0.77 & 56.6 & 0.32 \\
5,000 & 98.4 & 0.17 & 0.78 & 56.8 & 0.33 \\
10,000 & 98.5 & 0.16 & 0.79 & 57.3 & 0.29 \\
\bottomrule
\end{tabular}
\caption{The performance of \ourmodel with varying candidate number $N$ during infilling.}
\label{tab:effect_candidate_num}
\end{table}

\begin{table}[htbp]
\centering
\begin{tabular}{@{} l ccccc @{}}
\toprule
$d$ & \textbf{ASR}$\uparrow$ & \textbf{DIS}$\downarrow$ & \textbf{BLEU}$\uparrow$ & \textbf{PPL}$\downarrow$ & \textbf{GER}$\downarrow$ \\
\midrule
2 & 27.4 & 0.16 & 0.79 & 60.6 & 0.33 \\
4 & 98.4 & 0.17 & 0.78 & 56.8 & 0.33 \\
6 & 98.8 & 0.18 & 0.78 & 55.7 & 0.25 \\
10 & 98.5 & 0.20 & 0.76 & 55.9 & 0.23 \\
25 & 98.7 & 0.22 & 0.74 & 55.8 & 0.19 \\
\bottomrule
\end{tabular}
\caption{The performance of \ourmodel with varying depth restriction $d$ when selecting phrase candidates.}
\label{tab:effect_depth}
\end{table}

\begin{table}[htbp]
\centering
\begin{tabular}{@{} l ccccc @{}}
\toprule
$l$ & \textbf{ASR}$\uparrow$ & \textbf{DIS}$\downarrow$ & \textbf{BLEU}$\uparrow$ & \textbf{PPL}$\downarrow$ & \textbf{GER}$\downarrow$ \\
\midrule
2 & 98.4 & 0.17 & 0.78 & 58.5 & 0.32 \\
3 & 98.4 & 0.17 & 0.78 & 56.8 & 0.33 \\
6 & 98.6 & 0.19 & 0.77 & 54.2 & 0.37 \\
10 & 99.2 & 0.21 & 0.76 & 55.4 & 0.45 \\
15 & 99.0 & 0.22 & 0.75 & 52.0 & 0.42 \\
\bottomrule
\end{tabular}
\caption{The performance of \ourmodel with varying substitution length incremental restriction $l$ when selecting phrase candidates.}
\label{tab:effect_length}
\end{table}

\section{Details of Human Evaluation}
\label{app:human_evaluation_details}

\begin{figure}[h]
\setlength{\belowcaptionskip}{-0.2cm}
    \centering
    \includegraphics[width=\linewidth]{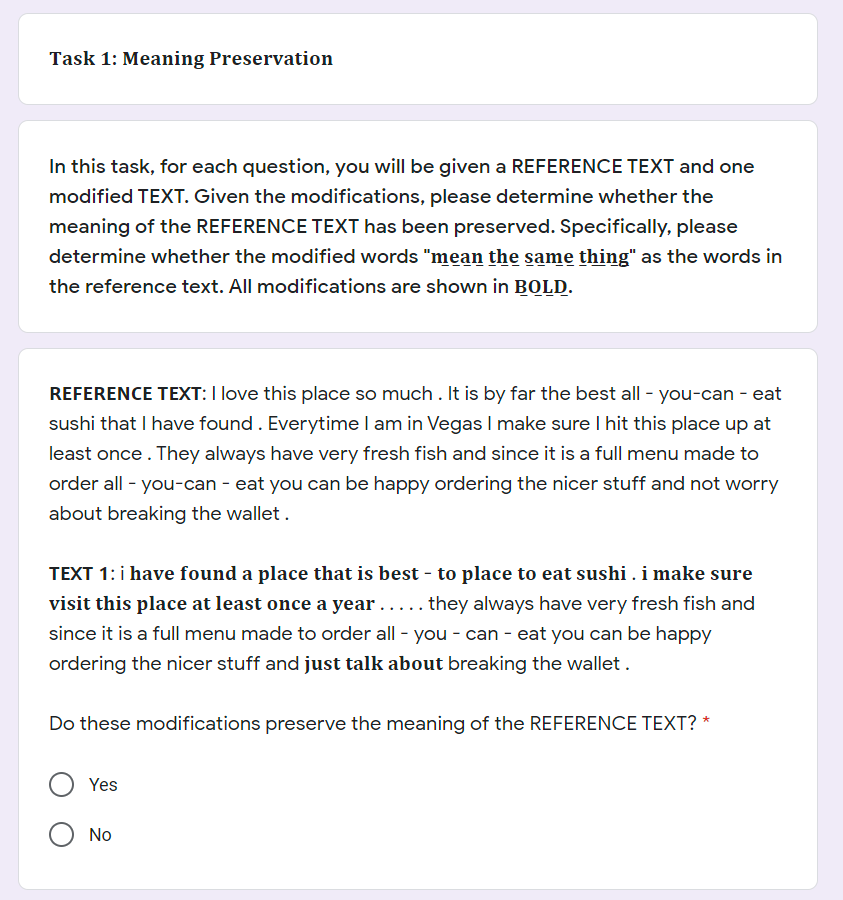}
    \caption{The instruction and an example of meaning preservation task in human evaluation.}
    \label{fig:human_meaning_preservation}
\end{figure}

\begin{figure}[t!]
\setlength{\belowcaptionskip}{-0.2cm}
    \centering
    \includegraphics[width=\linewidth]{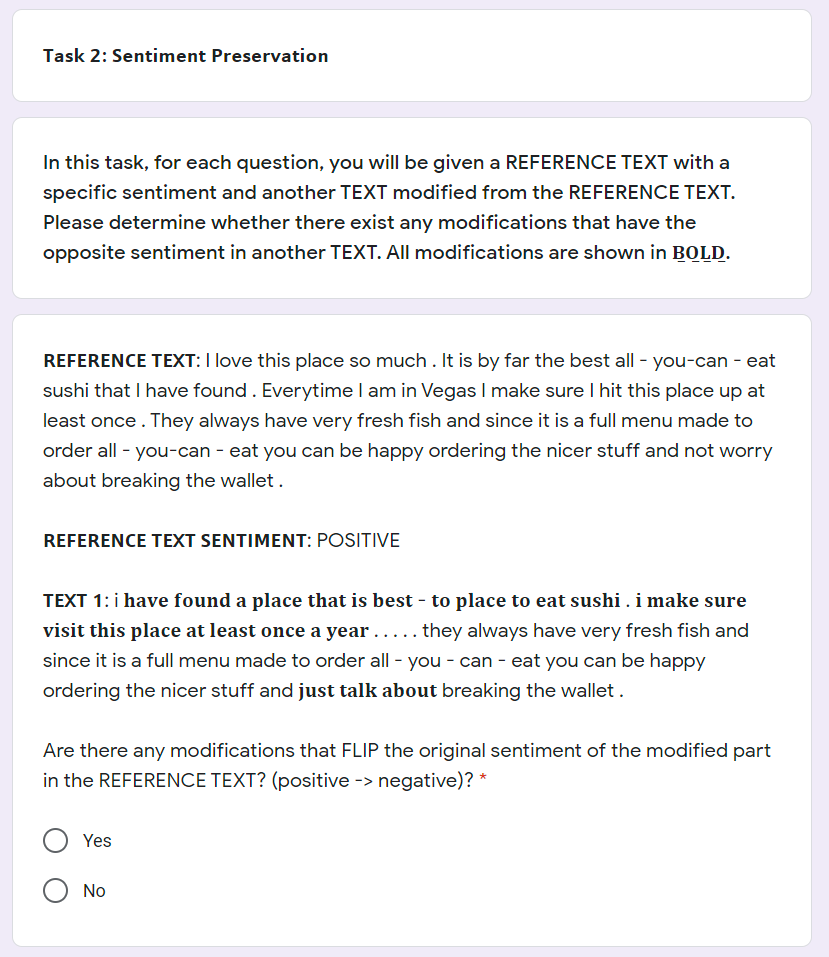}
    \caption{The instruction and an example of label preservation task in human evaluation.}
    \label{fig:human_label_preservation}
\end{figure}

\begin{figure}[t!]
\setlength{\belowcaptionskip}{-0.2cm}
    \centering
    \includegraphics[width=\linewidth]{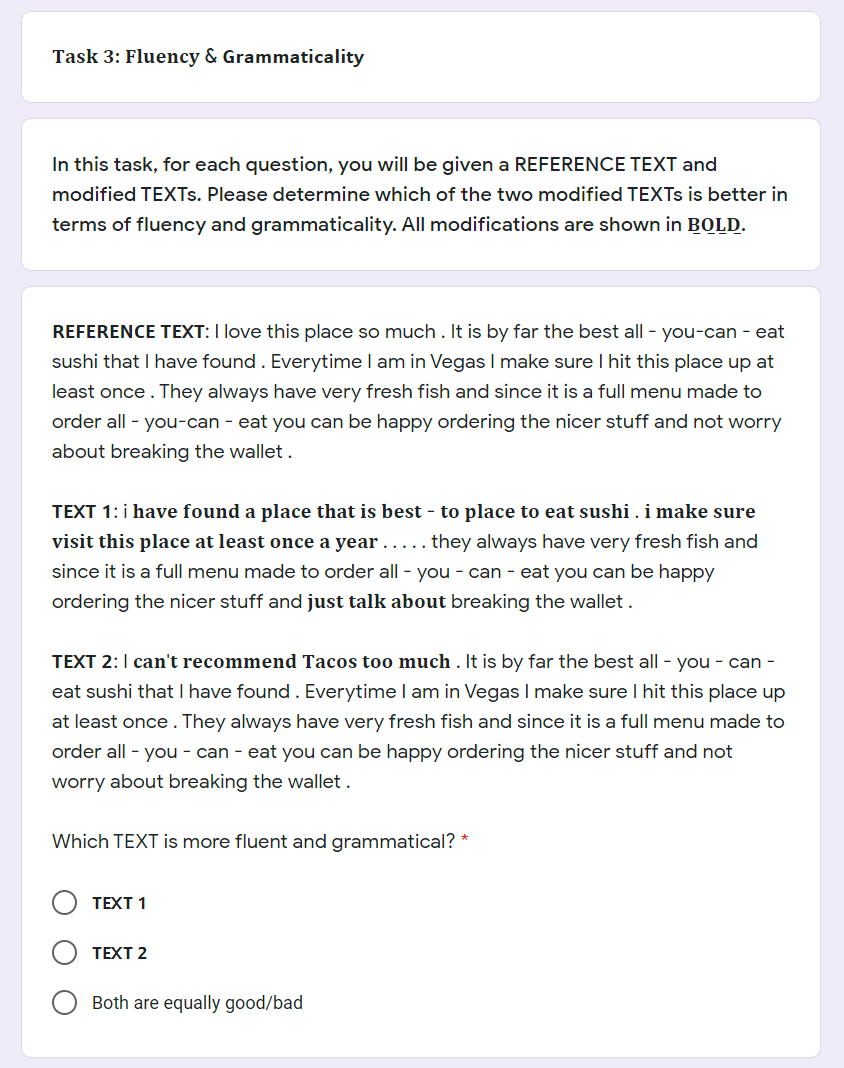}
    \caption{The instruction and an example of fluency and grammaticality comparison task in human evaluation.}
    \label{fig:human_fluency}
\end{figure}

We conducted our human evaluation via \emph{Google Forms} on a total 60 non-expert volunteer annotators. Each annotator was asked to rate for 10 sets of examples, where each set contains one original sample and three corresponding adversarial samples generated by \ourmodel, CLARE, and MAYA respectively. We show the screenshot of our instructions and examples in Figure~\ref{fig:human_meaning_preservation}, Figure~\ref{fig:human_label_preservation}, and Figure~\ref{fig:human_fluency}, where the perturbed parts are in bold font. We described how we would use these collected data in the invitations for annotators, and they must agree on this usage before evaluation. All collected data go without personal information in our experiments.

\end{document}